\documentclass[letterpaper]{article} 
\usepackage{aaai2026}  
\usepackage{times}  
\usepackage{helvet}  
\usepackage{courier}  
\usepackage[hyphens]{url}  
\usepackage{graphicx} 
\usepackage{natbib}  
\usepackage{caption} 
\usepackage{algorithm}
\usepackage{algorithmic}

\usepackage{algorithm}
\usepackage{algorithmic}

\usepackage{amssymb}
\usepackage{booktabs}
\usepackage{threeparttable}
\usepackage{multirow}
\usepackage{colortbl}
\usepackage{xcolor}
\definecolor{dark-gray}{gray}{0.30}
\newcommand{\pub}[1]{{\color{dark-gray}{\scriptsize{[{#1}]}}}}
\usepackage{cleveref}
\usepackage{newfloat}
\usepackage{listings}
\DeclareCaptionStyle{ruled}{labelfont=normalfont,labelsep=colon,strut=off} 
\floatstyle{ruled}
\newfloat{listing}{tb}{lst}{}
\floatname{listing}{Listing}
\title{PriorDrive: \\Enhancing Online HD Mapping with Unified Vector Priors}
\author{
Shuang Zeng\textsuperscript{\rm 1,2*\dag}, Xinyuan Chang\textsuperscript{\rm 2\dag}, Xinran Liu\textsuperscript{\rm 2}, \\ Yujian Yuan\textsuperscript{\rm 3}, Shiyi Liang\textsuperscript{\rm 1,2}, Zheng Pan\textsuperscript{\rm 2}, Mu Xu\textsuperscript{\rm 2},  Xing Wei\textsuperscript{\rm 1\ddag}
}

\affiliations{
     \textsuperscript{\rm 1}Xi’an Jiaotong University,
 \textsuperscript{\rm 2}Amap, Alibaba Group,
  \textsuperscript{\rm 3}The Hong Kong University of Science and Technology \\
\tt\small \{zengshuang, sy\_liang2023\}@stu.xjtu.edu.cn, weixing@mail.xjtu.edu.cn, yyuanbn@connect.ust.hk \\
\tt\small \{changxinyuan.cxy, tom.lxr, panzheng.pan, xumu.xm\}@alibaba-inc.com
}

\usepackage{bibentry}

\begin{document}

\twocolumn[{%
	\renewcommand
	\twocolumn[1][]{#1}%
	\maketitle
		\centering
		\vspace{-15pt}
            \includegraphics[width=0.95\textwidth]{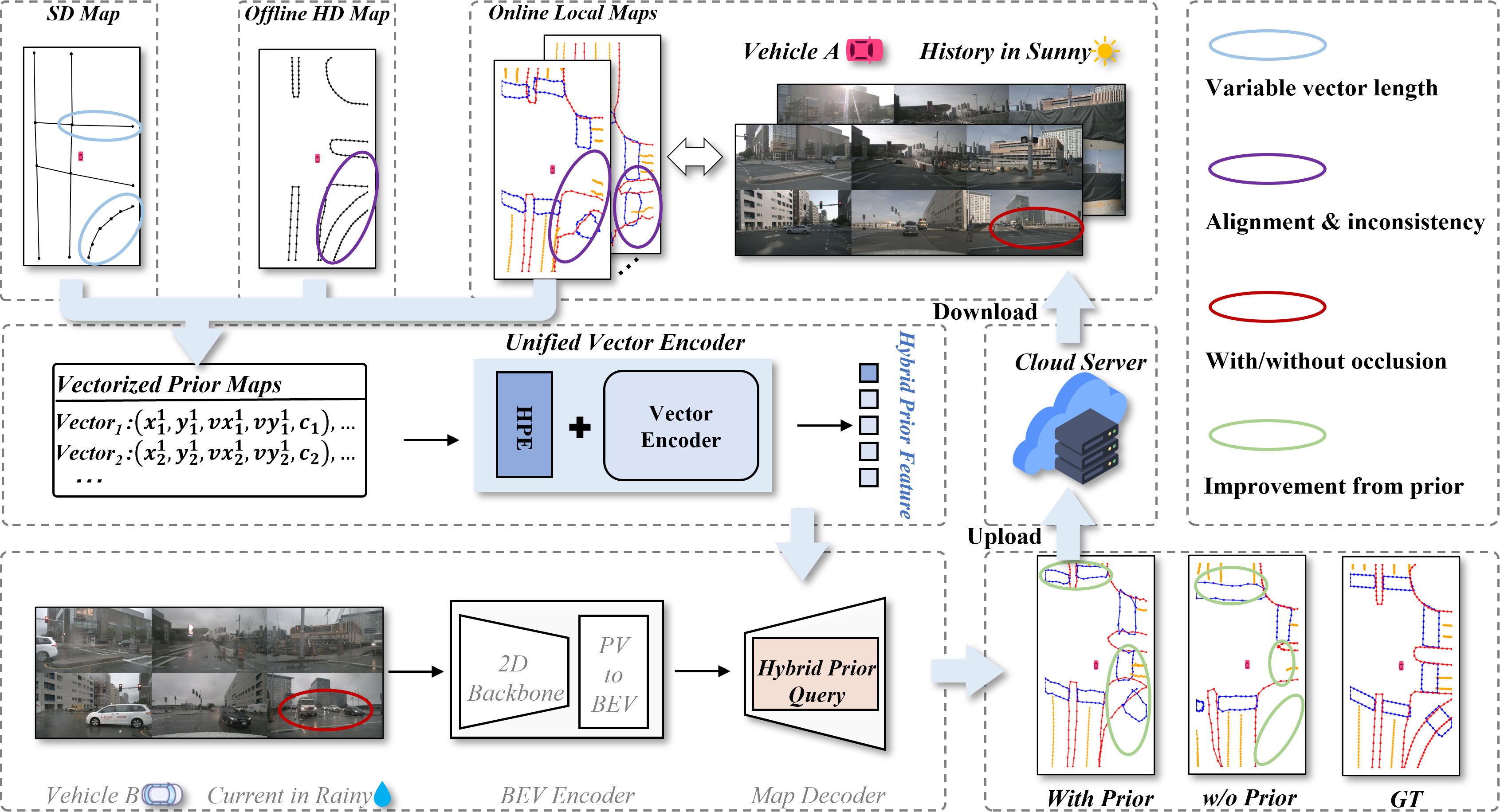}
\captionof{figure}{Overview of PriorDrive. PriorDrive seamlessly integrates diverse vectorized prior maps into existing online mapping frameworks, making final predictions more complete and accurate than those generated without priors. The optimized predictions can be uploaded to cloud servers for other vehicles to download and use as prior maps.}
\label{fig:priors}
}]

\renewcommand\thefootnote{}\footnotetext{$^*$ Work done during the internship at Amap, Alibaba Group.}
\renewcommand\thefootnote{}\footnotetext{$^\dag$ Equal contribution.}
\renewcommand\thefootnote{}\footnotetext{$^\ddag$Corresponding author. This work was supported by the National Natural Science Foundation of China No. 62572385 and the Fundamental Research Funds for the Central Universities No. xxj032023020.}

\begin{abstract}
High-Definition Maps (HD maps) are essential for the precise navigation and decision-making of autonomous vehicles, yet their creation and upkeep present significant cost and timeliness challenges. The online construction of HD maps using on-board sensors has emerged as a promising solution; however, these methods can be impeded by incomplete data due to occlusions and inclement weather, while their performance in distant regions remains unsatisfying. This paper proposes PriorDrive to address these limitations by directly harnessing the power of various vectorized prior maps, significantly enhancing the robustness and accuracy of online HD map construction.
Our approach integrates a variety of prior maps uniformly, such as OpenStreetMap's Standard Definition Maps (SD maps), outdated HD maps from vendors, and locally constructed maps from historical vehicle data. To effectively integrate such prior information into online mapping models, we introduce a Hybrid Prior Representation (HPQuery) that standardizes the representation of diverse map elements.
We further propose a Unified Vector Encoder (UVE), which employs fused prior embedding and a dual encoding mechanism to encode vector data. To improve the UVE's generalizability and performance, we propose a segment-level and point-level pre-training strategy that enables the UVE to learn the prior distribution of vector data.
Through extensive testing on the nuScenes, Argoverse 2 and OpenLane-V2, we demonstrate that PriorDrive is highly compatible with various online mapping models and substantially improves map prediction capabilities. The integration of prior maps through PriorDrive offers a robust solution to the challenges of single-perception data, paving the way for more reliable autonomous vehicle navigation. Code is available at https://github.com/MIV-XJTU/PriorDrive
\end{abstract}

\section{Introduction}
\label{sec:intro}
High-Definition Maps (HD maps) are essential for the accurate navigation and decision-making of autonomous vehicles, providing detailed vectorized representations of road elements~\cite{HDmaps-survey,Gao2020VectorNetEH,yao2024waterscenes,yao2025exploring,lu2025differentiable}. Despite their significance, the traditional methods for creating and maintaining HD maps are often costly and labor-intensive. These methods can result in outdated maps that struggle to keep pace with the rapidly changing urban environments. In response, there is increasing interest in online HD map construction~\cite{BEMAPNET,PivotNet,MAPVR,yao2025usvtrack}, where maps are generated in real-time using on-board sensors. Although this approach reduces costs, it also introduces new challenges, particularly concerning incomplete and error-prone data caused by environmental occlusions (see red circle in~\Cref{fig:priors}) or inclement weather conditions, coupled with unsatisfactory performance in distant regions.

Using prior maps to alleviate the problem of online mapping is a promising solution. Specifically, there are three common types of prior maps: 1) Standard Definition Maps (SD maps), 2) existing offline outdated HD maps, and 3) online historical prediction local maps, as illustrated in the upper left corner of \Cref{fig:priors}. SD maps provide crucial long-distance centerline skeletons, but with uncertain accuracy compared to HD maps. While existing HD maps (HD map-EX) offer high accuracy, their infrequent updates may lead to outdated information that fails to fully reflect current road conditions. Historical predicted local maps can offer insights by incorporating previous observations, but being a single mapping result, they may not guarantee the completeness and accuracy of the prior map.

Generally, there are two ways to encode vectors: 1) rasterized, 2) vectorized. Recent studies~\cite{Xiong2023NeuralMP,Jiang2024PMapNetFM,li2024towards1,li2025hyfacialhybrid} have attempted to use rasterized or vectorized prior maps to alleviate the problem of online mapping. However, existing prior-based methods~\cite{Xiong2023NeuralMP,Jiang2024PMapNetFM,MAPEX,10.1117/12.3049486,li2025lion} face several limitations. For instance, P-MapNet~\cite{Jiang2024PMapNetFM}, NMP~\cite{Xiong2023NeuralMP} and HRMapNet~\cite{zhang2024hrmapnet} leverage rasterized SD maps or historical prediction maps. The rasterized representation, limited by resolution, is lossy and redundant while lacking the detail necessary to capture vectorized instance-level information and express the type and direction of map elements effectively. Furthermore, P-MapNet and HRMapNet require complex post-processing to convert vectorized maps into rasterized maps, which also do not align with the native vector storage format of maps. On the other hand, MapEX~\cite{MAPEX} exclusively relies on vectorized outdated HD maps, which struggle to reflect real-time changes in road structures, due to low update frequencies, making them unreliable for current navigation needs. Moreover, the encoding scheme of MapEX is overly simplistic, thereby failing to effectively capture instance-level information within vector representations.

It is worth mentioning that encoding vectors is not a straightforward task. There are several main challenges: 1) Different types of vector maps may contain varying vector types (e.g., points, lines, planes), 2) Vector lengths can be fixed or variable (see blue circle in~\Cref{fig:priors}), and 3) Alignment and inconsistency issues exist between different vector maps (see purple circle in~\Cref{fig:priors}). 
Previous methods were limited to efficiently encoding only a single type of prior map, but the complementary information contained across different prior map categories. For instance, NavMap~\cite{schmidt2023navmap} omitted road segments with variable lengths when integrating HD maps and SD maps, resulting in the loss of valuable information. Furthermore, previous methods generally encode only vector positions and categorical attributes, lacking the capability to capture finer-grained information such as direction, geometric shape, and topology.

In this paper, we overcome these challenges and introduce \textbf{PriorDrive}, a novel framework designed to directly and unifiedly integrate diverse vectorized prior maps into various mapping models, fully leveraging their respective strengths and complementary information. We propose a Unified Vector Encoder (UVE) that encodes various vector data and captures fine-grained vector information. It can extract fixed-length instance-level and point-level features from variable-length vectors. 
Additionally, we introduce for the first time a point-level and segment-level pre-training paradigm specifically tailored for vector data, aiming to enhance the model's understanding and representation of vector.
Concurrently, we propose a hybrid prior representation (HPQuery) to represent all elements, facilitating the integration of unified vector features into mapping models. Experiments on nuScenes, Argoverse 2 and OpenLane-V2 demonstrate that PriorDrive is plug-and-play, seamlessly applicable to diverse online mapping models and vector tasks with significant performance improvements.

In summary, our contributions are as follows:
\begin{itemize}
\item We introduce a unified vector encoder that effectively captures diverse vector attributes and enables encoding of variable-length vectors into fixed-length features through fused prior embedding and dual encoding mechanism.  
\item We first proposed a point-level and segment-level pre-training paradigm for vector data, effectively capturing prior distribution and mitigating noise in historical maps.
\item We propose a hybrid prior representation (HPQuery) to represent all elements and a PriorDrive framework with various vector prior maps to address the limitations of single-perception. Our comprehensive evaluation demonstrates that PriorDrive is plug-and-play and  significantly enhances the online mapping models.
\end{itemize}

\section{Related Work}
\label{sec:related}

\subsection{Online Vectorized HD Map Construction}
Online vectorized HD map construction was initially treated as a segmentation task~\cite{bevformer,zhang2024self,sun2023large,guo2025depth,cui2025efficient}. To construct a vectorized HD map, HDMapNet~\cite{Li2021HDMapNetAO} requires complex post-processing of pixel-level rasterized maps. VectorMapNet~\cite{Liu2022VectorMapNetEV} introduced a coarse-to-fine, two-stage network leveraging keypoint representations. MapTR~\cite{liao2022maptr} advanced this concept by modeling point sets in a permutation-equivalent manner using a DETR-like~\cite{detr,9681188,zeng2025janusvln,xiao2025worldenv,jin2025reasoning} one-stage network. Subsequent studies~\cite{HIMAP,sun2025yolov4svm,zeng2024driving,lan2025contextual,lan2025mappo} have introduced various insightful approaches for further improvement. However, these methods rely solely on single-source perception data from vehicle-mounted sensors, limiting their effectiveness in challenging scenarios such as occlusion, inclement weather conditions or distant regions.

The release of OpenLane-V2~\cite{wang2023openlanev2} dataset has sparked growing interest in topology reasoning within driving scenes. This task focuses on recognizing the topologies among lanes and lanes with traffic elements. TopoNet~\cite{li2023toponet} leverages GNN to update lane representation and lane topology. TopoMLP~\cite{wu2023topomlp} utilizes PETR~\cite{Liu2022PETRPE} for centerline detection and employs simple MLPs to predict the relationships. In this paper, we also apply PriorDrive to topological inference tasks, demonstrating that PriorDrive is plug and play and widely applicable to other vector tasks.

\subsection{Online Mapping Based on Prior Maps}
Recent research has explored the integration of prior maps to enhance the performance of online mapping models. P-MapNet~\cite{Jiang2024PMapNetFM} encodes Standard Definition Maps (SD map) as an additional conditional branch and employs a masked autoencoder to capture the prior distribution of HD maps. NMP~\cite{Xiong2023NeuralMP} introduces a global neural map prior that self-updates, thereby improving the performance of local map inference. MapEX~\cite{MAPEX} utilizes existing HD maps and refines the query-based map estimation model’s matching algorithm. Although these advancements have achieved considerable results, many studies have overlooked the potential of online historical prediction maps as a source of prior information. Moreover, previous methods~\cite{zhang2024hrmapnet,Ma2024RoadPainterPA,wan2025driving,yuan2025unimapgen,li2025multi} typically could only utilize a single type of prior map. Our work can uniformly encode various types of vector prior maps to enhance current perception data, fully leveraging the complementary advantages among different prior maps.

\subsection{Pretrained Methods Based on Mask Modeling}
In the fields of NLP and CV, masked pre-training has proven to be an effective strategy for self-supervised representation learning. In NLP, models like BERT~\cite{betr,Liu2019RoBERTaAR,Lan2019ALBERTAL,zeng2025FSDrive,li2023ultrare} use masked language modeling to predict randomly masked tokens within a bidirectional text context. Similarly, methods such as MAE~\cite{MAE,Liu2022MixMAEMA,liang2025persistent,dai2025unbiased,yin2025knowledge} mask random patches of input images and reconstruct them based on the remaining unmasked patches. Diverging from these approaches~\cite{zhang2025molebridge,ZHANG2025103888,wei2025copeft,wei2026infocom,li2025cogvla}, which are predominantly tailored for text and image data, we introduce a pre-training paradigm specifically designed for vector data. Our method aims to learn the prior distribution of vector representations, offering a new avenue for pre-training in the context of vectorized data.

\section{Proposed Method: PriorDrive}
\label{sec:method}


\subsection{Problem Formulation}
\label{sec:formulation}

\paragraph{Formulation of Online Mapping.}
The goal of online mapping is to construct a local HD map using on-board sensor observations, such as the surrounding camera images. The online mapping process typically involves two main modules: the Bird’s-Eye View (BEV) encoder and the map decoder. The BEV encoder generally includes a Perspective View (PV) feature extractor and a module for transforming PV features into BEV features. The map decoder usually comprises a Fully Convolutional Network~\cite{Shelhamer2014FullyCN,hu2024tmff,202508.0462,huang2024ar} or a query-based module similar to DETR~\cite{detr,xie2025seqgrowgraph,tang2025dissecting,tang2025ocrt}, which outputs either rasterized perception results or vectorized map elements.
\begin{figure}[t!]
\centering
\includegraphics[width=\columnwidth]{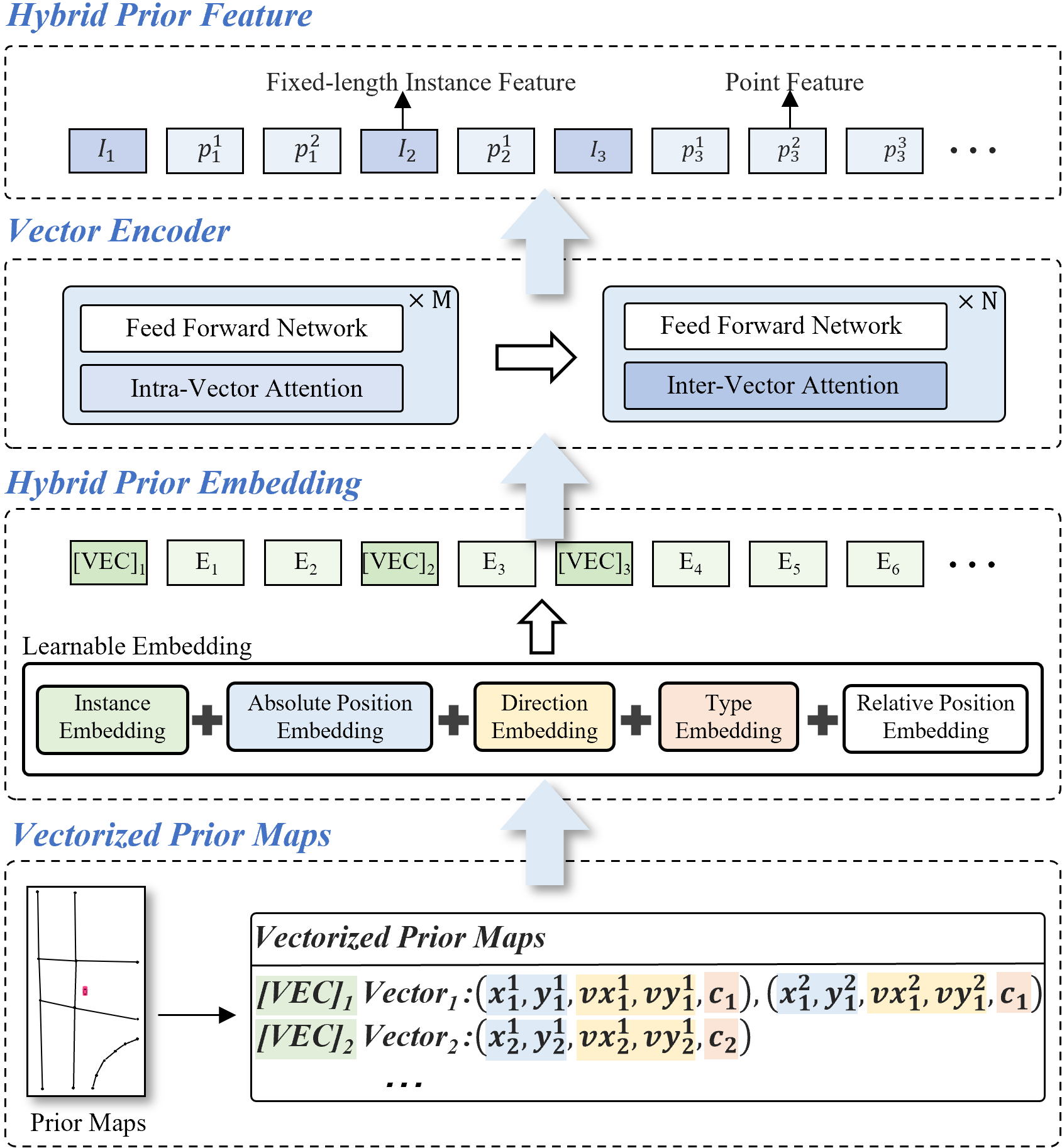}
\caption{Structure of UVE. Fused prior embedding enhances vector representation. Intra-instance and inter-instance attention refine local features and capture global context effectively.}
\label{fig:UVE}
\end{figure}

Let the set of images be denoted as $\{I_{1},\ldots,I_{k}\}$, and the BEV encoder be represented as ${E}_{bev}$. The extraction of BEV features ${f}_{bev}$ can then be expressed as:
{\setlength{\abovedisplayskip}{5pt}
 \setlength{\belowdisplayskip}{5pt}
 \begin{equation}
  f_{bev}={E}_{bev}(\{I_1,\ldots,I_K\}).
 \end{equation}
}

Similarly, let the map decoder module be denoted as $D_{map}$. The final process of predicting map elements can be expressed as:
{\setlength{\abovedisplayskip}{5pt}
 \setlength{\belowdisplayskip}{5pt}
 \begin{equation}
  \mathcal{P}={D}_{map}(f_{bev}, ~ opt({Q})),
 \end{equation}
}
where $\mathcal{P}$ represents the predicted map elements, $Q$ denotes the learnable query, and $opt$ indicates that this learnable query is optional, which is required in query-based models but not in other models.

\subsection{Architecture of UVE}
\label{sec:uve}

Inspired by BERT~\cite{betr,he2025social,rs17132267,ren2025digital} on text-related tasks, we propose a unified vector encoder (UVE) by analogizing vector points to words and vector elements to sentences (see~\Cref{fig:UVE}).

\paragraph{Extraction of Prior Map Features.}
Our proposed UVE serves as a unified vector encoder that can directly encode various vector data information, such as the position ($x$, $y$), direction ($v_x$, $v_y$), and type of points ($c$). Using $p=\left[x, y, v_x, v_y, c\right]$ to represent the point, a vector $v$ typically consists of an ordered set of a variable number of points, $v=\left[p_1, p_2, \ldots, p_n\right]$, where $n$ varies. The map prior $M$ consists of a set of vectors $M=\{v_1, v_2, \ldots, v_m\}$, with $m$ being variable. When employing different prior maps, all of prior maps were integrated into a cohesive whole, $M_{prior}=\{M_1, M_2, \ldots, M_t\}$. Using $E_{uve}$ to represent the UVE model, the extraction process of the features of the prior map, denoted as $f_{prior}$, can be represented as:
{\setlength{\abovedisplayskip}{5pt}
 \setlength{\belowdisplayskip}{-5pt}
 \begin{equation}
  f_{prior}=E_{uve}(M_{prior}).
 \end{equation}
}

\paragraph{Fused Prior Embedding.} 
As shown in~\Cref{fig:UVE}, for the given prior maps $M_{prior}$, we first construct the input for UVE, termed as Fused Prior Embedding (FPE). We use the method from~\cite{TancikSMFRSRBN20,,zeng2025lens,lin2025plan,yang2025unleashing} for obtaining point position embedding and direction embedding for both $(x, y)$ and $(v_x, v_y)$. We concatenate these two embeddings as the point-level embedding. 
To obtain the fixed-length feature of the entire vector, we add a special \emph{[VEC]} token at the beginning of each vector and use the embedding of this token as the instance-level embedding.
Additionally, we introduce learnable instance embedding and learnable type embedding to distinguish vector instances. To ensure the order of points, we also introduce 2D learnable position embedding. Finally, these embeddings are aggregated to form FPE.

\begin{figure}[t!]
\centering
\includegraphics[width=\columnwidth]{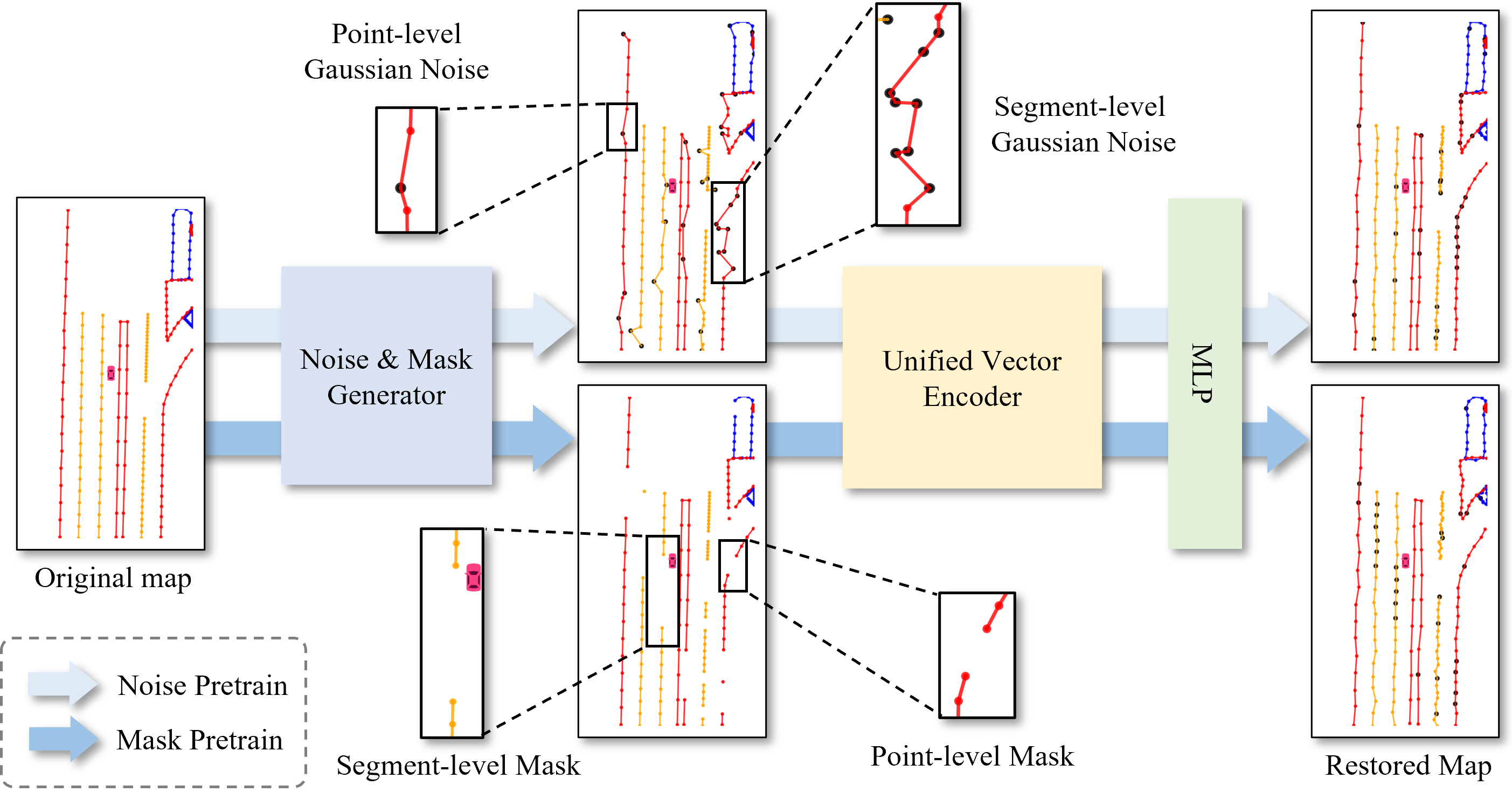}
\caption{Pre-trained Pipeline. The noise \& mask generator creates segment-level and point-level noise or mask, and then reconstructs the entire map via UVE and MLP.}
\label{fig:pretrain}

\end{figure}

\paragraph{Dual Encoding Mechanism.} 
Due to the limited information content of vector points themselves, learning the concept of vector instances is very challenging, requiring more interactions among the points within the vectors to enhance each point's perceptual abilities towards others in the same vector. Therefore, UVE encoder utilizes $M$-layer intra-vector attention and $N$-layer inter-vector attention, using an attention masking mechanism to facilitate feature interactions within and between different vector instances. Each mask in the intra-vector encoder has a value of 1 at its own position and 0 elsewhere, while the inter-vector encoder employs a mask matrix filled entirely with 1s.

When using different prior maps and multiple online local maps, there are 
alignment problems between different prior maps and potential inconsistencies between online local maps generated across different trips or different timestamps in one trip. Our UVE mitigated this challenge by employing fused prior embedding and dual encoding mechanism to enhance the model's ability to learn instances, supporting adaptive focus on the superior map elements in different prior maps. The experiments presented in~\Cref{tab:abla} substantiate this capability.

\subsection{Pre-training UVE: Position Modeling}
\label{sec:pre-train}
Due to the limited inference ability of online mapping models, there are errors in historical prediction maps. So we used position modeling to pre-train UVE to improve its encoding and noise reduction capabilities. Specifically, there are two methods: random noise and masking (see \Cref{fig:pretrain}). 
\paragraph{Noise \& Mask Generator.} 
The noise is primarily categorized into segment-level and point-level.
In point-level noise, random noise is added to $5\%$ of the vector points across the entire map. For segment-level noise, random noise is added to all vector points within a sub-vector segment, after randomly selecting $10\%$ of the map elements. Using $M_{org}$ to represent the original map elements (with the same data structure as $M$), the process of adding Gaussian noise can be represented by the following formula:
{\setlength{\abovedisplayskip}{5pt}
 \setlength{\belowdisplayskip}{7pt}
 \begin{equation}
  M_{org}^* = M_{org}\emph{[random\_index]} + \epsilon,
 \end{equation}
}
where $\epsilon \sim \mathcal{N}(0, 1)$ is a random noise, which is only added to the horizontal and vertical coordinates of each point.

Similar to adding noise, the selection of points for adding the mask is also divided into point-level and segment-level. After selecting the points to add the mask, the coordinates of these points will be masked as $-1$.
{\setlength{\abovedisplayskip}{5pt}
 \setlength{\belowdisplayskip}{0pt}
 \begin{equation}
 M_{org}\emph{[random\_index]} = Mask.
 \end{equation}
}
\paragraph{Loss Function.} We feed vector map to UVE encoding after passing through the noise mask generator, then decode the coordinates of all points using MLP, and calculate mean euclide distance from GT coordinates as supervision:
{\setlength{\abovedisplayskip}{5pt}
 \setlength{\belowdisplayskip}{5pt}
 \begin{equation}
 \mathcal{L}= RMSE( P , ~ mlp(E_{uve}(M_{org}^*))),
 \end{equation}
}
where $P=\{(x_i, y_i)\}_{i=0}^{k}$ represent all the points in $M_{org}$, $RMSE$ on behalf of the root mean square error.

\begin{figure}[!t]
\centering
\includegraphics[width=\columnwidth]{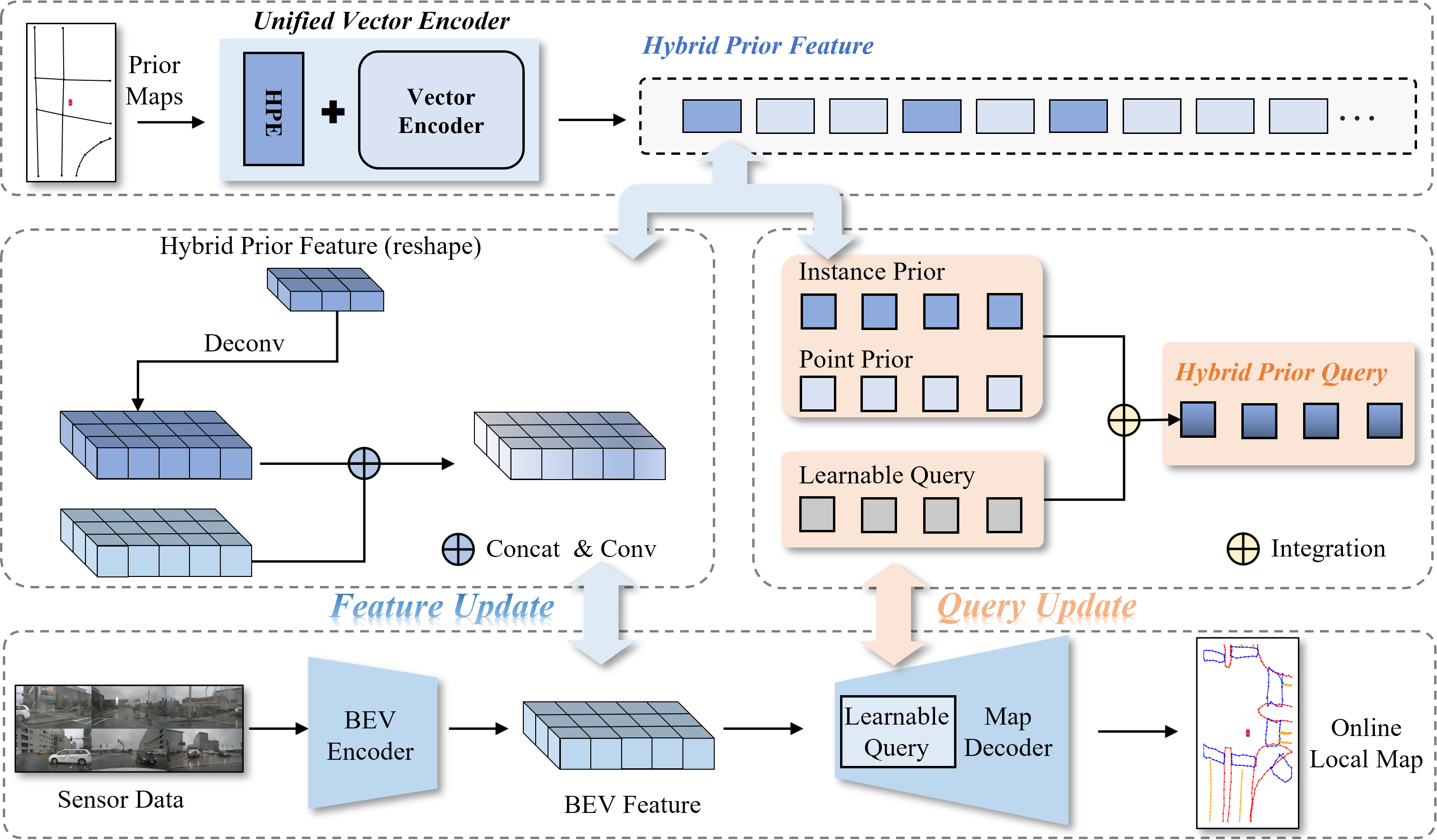}
\caption{Our Integration Method. For query-based models, we enable interaction between the prior features and queries at both instance-level and point-level, ensuring a more nuanced integration. For the other models, we integrate prior features directly with the BEV features.} 
\label{fig:method}

\end{figure}

\subsection{Integrated Approaches to BEV and HPQuery}
\label{sec:hpquery}
We respectively proposed different integration methods for the query-based/non-query-based approaches to integrate the prior map features into the online map construction model. For methods without a learnable query $Q$, such as HDMapNet~\cite{Li2021HDMapNetAO}, we directly reshape the prior feature $f_{prior}$ and use $Deconv$ upsampling to match the shape of the BEV feature $f_{bev}$. Then connect it with the BEV feature $f_{bev}$ and aligned by $Conv$, as shown in \Cref{fig:method} left.
{\setlength{\abovedisplayskip}{7pt}
 \setlength{\belowdisplayskip}{7pt}
 \begin{equation}
 f_{bev}^*=Conv(cat[f_{bev}, ~ Deconv(reshape(f_{prior})) ]),
 \end{equation}
}

here, $f_{bev}^*$ represents BEV feature after incorporating the prior map information, which can be input into the map decoder $D_{map}$ to obtain the predicted results of map elements.

For methods based on learnable query $Q$, such as the vectorized model MapTR series~\cite{maptrv2}, we propose HPQuery of three operations: addition, replacement, and concatenation, as shown in \Cref{fig:method} right. Specifically, $Q$ is typically composed of two parts, $Q = \{q_{ij}\}_{i=0,j=0}^{m,n} = \{q_i^{ins} + q_j^{pt}\}_{i=0,j=0}^{m,n}$, $q_i^{ins}$ and $q_i^{pt}$ representing the learnable queries at the instance-level and point-level respectively. The features $f_{prior}$ of prior maps are also composed of instance-level features and point-level features, $f_{prior} = \{(f_i^{ins}, f_j^{pt})\}_{i=0,j=0}^{m',n'}$. So we interact prior features and queries at the instance-level and point-level, respectively. The HPQuery of addition, replacement, and concatenation operations can be expressed as:

\begin{equation}
    q_{ij}^*=q_i^{ins}.add(f_i^{ins}) + q_j^{pt}.add(f_j^{pt}),
\end{equation}
\begin{equation}
    q_{ij}^*=q_i^{ins}.replace(f_i^{ins}) + q_j^{pt}.replace(f_j^{pt}),
\end{equation}
\begin{equation}
    q_{ij}^*=concat[q_i^{ins}, ~ f_i^{ins}] + concat[q_j^{pt}, ~ f_j^{pt}].
\end{equation}

We will detail the effectiveness of these feature fusion methods in supplementary material.

\section{Experiments}
\label{sec:experiments}
\subsection{Experimental Settings}

\paragraph{Benchmark.} We mainly validate our method on the popular nuScenes~\cite{Caesar2019nuScenesAM} and Argoverse 2~\cite{Argoverse2} following the previous methods~\cite{zhang2024hrmapnet,maptrv2,zhou2025hademif,zhou2025valuing}. NuScenes contains a total of 1,000 scenes, which was collected utilizing a 32-beam LiDAR operating and six cameras offering a 360-degree field of view. Argoverse 2 includes 1000 logs, which was captured from 7 cameras, along with a 3D vector map.

To demonstrate PriorDrive is plug-and-play and widely applicable to other vector tasks, we conducted topology inference on OpenLane-V2~\cite{wang2023openlanev2} Subset A.

\paragraph{Evaluation Protocol.} For vectorized results, we follow the standard metric used in previous works~\cite{Liu2022VectorMapNetEV,maptrv2,li2025revolutionizing,li2025knowledge}. The perception range along the direction of vehicle travel is $[-15.0m,15.0m]$ for $X$-axis and $[-30.0m, 30.0m]$ for $Y$-axis. We compute the average precision (AP) $\mathrm{AP}_{\tau}$ under several Chamfer distance $D_{Chamfer}$ thresholds $(\tau\in T,T=\{0.5,1.0,1.5\})$, and then calculate the mean across all thresholds as the final AP metric:
{\setlength{\abovedisplayskip}{5pt}
 \setlength{\belowdisplayskip}{5pt}
 \begin{equation}
\mathrm{AP}=\frac1{|\mathrm{T}|}\sum_{\tau\in\mathrm{T}}\mathrm{AP}_{\tau}.
 \end{equation}
}

\begin{table*}[t]\small
    \centering
    \begin{threeparttable}
    \setlength{\tabcolsep}{2.0mm}
    \begin{tabular}{l|c|ccc|cccc}
    \toprule
     \textbf{Method}& \textbf{Prior Map} & \textbf{Backbone} & \textbf{Epoch}&\textbf{Metric} & \textbf{Ped.} & \textbf{Div.} & \textbf{Bou.} & \cellcolor{gray!30}\textbf{Mean}\\

    \midrule
    
     {HDMapNet~\pub{ICRA21}~\cite{Li2021HDMapNetAO}}&{$\times$} &{Effi-B0} & {30} &{IoU}& {18.7} & {40.6} & {39.5} &  \cellcolor{gray!30}{32.9}  \\
      {P-MapNet~\pub{RAL24}~\cite{Jiang2024PMapNetFM}} & {$\checkmark$} &  {Effi-B0} &  {30} & {IoU}& 22.6 & 44.1 & 43.8 & \cellcolor{gray!30}36.8\\ 
      {NMP~\pub{CVPR23}~\cite{Xiong2023NeuralMP}} & {$\checkmark$} &  {Effi-B0} &  {30} & {IoU}& 21.0 & 44.2 & 46.1 & \cellcolor{gray!30}37.1\\ 
      {PriorDrive (Ours)} &  {$\checkmark$}&  {Effi-B0} &  {30} & {IoU}  & \textbf{25.1} & \textbf{46.7} & \textbf{48.4} & \cellcolor{gray!30}\textbf{40.1}
      \\ 
      \midrule
      {MapTR~\pub{ICLR23}~\cite{liao2022maptr}}& {$\times$}&  {R50} &  {24} & {AP}&  {41.2} &  {49.5} & {51.1} &  \cellcolor{gray!30}{47.3}  \\ 
      {P-MapNet~\pub{RAL24}~\cite{Jiang2024PMapNetFM}} & {$\checkmark$} &  {R50} &  {24} & {AP}& 43.7 & 50.9 & 53.5 & \cellcolor{gray!30}49.4\\    
      {StreamMapNet~\pub{WACV24}~\cite{Yuan_2024_streammapnet}} & {Temporal}&  {R50} &  {24} & {AP}& 60.4 & 61.9 & 58.9 &  \cellcolor{gray!30}60.4 \\  
      {PriorDrive (Ours)} &  {$\checkmark$}&  {R50} &  {24} & {AP}& \textbf{60.9}  & \textbf{65.3} & \textbf{63.8} & \cellcolor{gray!30}\textbf{63.3}\\  
      \midrule
      {MapTRv2~\pub{IJCV24}~\cite{maptrv2}}& {$\times$} &  {R50} &  {24} & {AP}&  {59.8} &  {62.4} & {62.4} &  \cellcolor{gray!30}{61.5}  \\
      {PrevPredMap~\pub{WACV25}~\cite{peng2024prevpredmap}} &   {$\checkmark$} &  {R50} &  {24} & {AP}& 64.5 & 66.9 & 67.6 &  \cellcolor{gray!30}66.3 \\
      {HRMapNet~\pub{ECCV24}~\cite{zhang2024hrmapnet}} &{$\checkmark$} &  {R50} &  {24} & {AP}& 65.8 & 67.4 & 68.5 &  \cellcolor{gray!30}67.2 \\
      {HRMapNet~\pub{ECCV24}~\cite{zhang2024hrmapnet}} &{$\checkmark$} &  {R50} &  {110} & {AP}& 72.0 & 72.9 & 75.8 &  \cellcolor{gray!30}73.6 \\
      {InteractionMap~\pub{CVPR25}~\cite{wu2025interactionmap}} &{Temporal} &  {R50} &  {24} & {AP}& 69.7 & 72.7 & 73.0 & \cellcolor{gray!30}71.8\\
      {InteractionMap~\pub{CVPR25}~\cite{wu2025interactionmap}} &{Temporal} &  {R50} &  {110} & {AP}& 75.3 & 75.6 & 77.0 & \cellcolor{gray!30}76.0\\
      {PriorDrive (Ours)} &{$\checkmark$} &  {R50} &  {24} & {AP}& \textbf{71.8} & \textbf{81.4} & \textbf{74.1} & \cellcolor{gray!30}\textbf{75.8}\\
      {PriorDrive (Ours)} &{$\checkmark$} &  {R50} &  {110} & {AP}& \textbf{78.0} & \textbf{82.0} & \textbf{81.6} & \cellcolor{gray!30}\textbf{80.5}\\
    
    \bottomrule
    \end{tabular}
    \end{threeparttable}
    \caption{Performance of various online mapping models with unified prior maps on nuScenes. All methods use camera input. In each block, the first row represents the baseline model, and all other methods are modifications upon the baseline. “Temporal” means using temporal information. The results of StreamMapNet and HRMapNet are from HRMapNet, while the other results are from the respective papers.} 
  
    \label{tab:main-result}
\end{table*}

\begin{table}[htbp]\small
\centering
    \begin{threeparttable}
    \setlength{\tabcolsep}{0.5mm}
    \begin{tabular}{lc|cccc}
    \toprule 
    \textbf{Method} &\textbf{Epoch}& $\textbf{AP}_{ped}$ & $\textbf{AP}_{div}$ & $\textbf{AP}_{bou}$ & \cellcolor{gray!30}\textbf{mAP}\\   
     \midrule
   MapTRv2~\pub{IJCV24} & 6& 60.7 & 68.9 & 64.5  &\cellcolor{gray!30}64.7\\
   HRMapNet~\pub{ECCV24} & 30& 65.1 & 71.4 & 68.6  &\cellcolor{gray!30}68.3\\
   PrevPredMap~\pub{WACV25} & 6& 64.1 & 71.4 & 67.4  &\cellcolor{gray!30}67.6\\
   InteractionMap~\pub{CVPR25}& 6& 66.6 & 75.6 & 72.7  &\cellcolor{gray!30}71.6\\
  {PriorDrive (Ours)} &6& \textbf{68.3} & \textbf{75.8} & \textbf{74.2} & \cellcolor{gray!30}\textbf{72.8}  \\

    \bottomrule
    \end{tabular}
    \end{threeparttable}
    \caption{Performance comparison of 3D map elements on Argoverse 2. }
  
    \label{tab:av2}
\end{table}

\begin{table*}[htbp]\small
\centering
    \begin{threeparttable}
    \begin{tabular}{l|c|cc|ccc|ccc}
    \toprule
      \multirow{2}{*}{\textbf{Method}}&  \multirow{2}{*}{\textbf{Prior Map}}&\multirow{2}{*}{\textbf{$\textbf{DET}_l$}} & \multirow{2}{*}{\textbf{$\textbf{DET}_t$}}&  \multicolumn{3}{c}{\textbf{v1.0.0}} &  \multicolumn{3}{c}{\textbf{v2.1.0}} \\
     & &&  & $\textbf{TOP}_{ll}$& $\textbf{TOP}_{lt}$&  \cellcolor{gray!30}\textbf{OLS} & $\textbf{TOP}_{ll}$& $\textbf{TOP}_{lt}$&  \cellcolor{gray!30}\textbf{OLS}  \\
     \midrule
   TopoNet + OSMR~\pub{IROS24}~\cite{osmr} &  {$\checkmark$}& 30.6 &44.6 & 7.7 & 22.9 & \cellcolor{gray!30}37.7& - & - & -\\ 

    RoadPainter~\pub{ECCV24}~\cite{Ma2024RoadPainterPA} &  {$\checkmark$}& \textbf{36.9}&  47.1& 12.7&25.8&\cellcolor{gray!30}42.6& -&-&- \\
    SMERF~\pub{ICRA24}~\cite{SMERF}&  {$\checkmark$} & 33.4& \textbf{48.6}& 7.5&23.4&\cellcolor{gray!30}39.4& 15.4&25.4&\cellcolor{gray!30}42.9 \\
 
   TopoLogic~\pub{NeurIPS24}~\cite{fu2024topologic}&  {$\times$} & 29.9 & 47.2 & 18.6 & 21.5 & \cellcolor{gray!30}41.6& 23.9 & 25.4 & \cellcolor{gray!30}44.1 \\
  {PriorDrive (Ours)} &  {$\checkmark$}&  31.5 & 48.4 & \textbf{24.6} &\textbf{30.7}  &\cellcolor{gray!30}\textbf{46.2} &\textbf{31.4}  & \textbf{33.7 }&\cellcolor{gray!30}\textbf{48.5}    \\
    \bottomrule
    \end{tabular}
    \end{threeparttable}
    \caption{Performance comparison with SOTA non-prior/prior-based methods on OpenLane-V2 subset\_A set. '-' denotes the absence of relevant data.}
    \label{tab:openlanev2}

\end{table*}

For the rasterized results, we follow the previous works~\cite{Xiong2023NeuralMP,202506.2087,chen2025visrl,chen2025sifthinker} and employ IoU as the metric for segmentation results.

For OpenLane-V2's lane centerline perception task, $DET_l$ averages the Frechet distance to quantify similarity. $DET_t$ uses IoU and averages it across various traffic categories. $TOP_{ll}$ and $TOP_{lt}$ compute the topology matrix similarity between lanes and between lanes and traffic elements, respectively. The overall metric is called OLS:
{\setlength{\abovedisplayskip}{5pt}
 \setlength{\belowdisplayskip}{5pt}
 \begin{equation}
\mathrm{OLS}=\frac{1}{4}[\mathrm{DET}_{l}+\mathrm{DET}_{t}+f(\mathrm{TOP}_{ll})+f(\mathrm{TOP}_{lt})],
 \end{equation}
}

where $f$ is the square root function.

\paragraph{Implementation Details.} 
For fair comparisons, experiments were conducted on all baseline models using their original hyperparameters. All experiments were trained with 8 NVIDIA RTX A6000. For pre-training UVE, we trained the 24-epoch on nuScenes vector data. This one-time offline pre-training (~12h) yields significant improvements, and its cost is amortized through plug-and-play reuse across various models. The sources of the three types of prior maps are as follows: The SD map is from OpenStreetMap~\cite{Haklay2008OpenStreetMapUS}; as for the existing HD prior maps, we follow MapEX~\cite{MAPEX}. For each map element localization, we add noise from a Gaussian distribution with a standard deviation of 1 meter. This has the effect of applying a uniform translation to each map element (dividers, boundaries, crosswalks); as for the predicted online local maps, in actual scenarios, the online local maps are generated by vehicles passing through the same area, and their data is stored on cloud servers for other vehicles to retrieve in the future. For datasets collected by a single vehicle such as nuScenes, Argoverse 2, and OpenLane-V2, certain road sections are captured through multiple drives. Therefore, we use each baseline model to infer the entire dataset, thereby generating online local prior maps at different trips or different times in the same trip, simulating this actual process. Due to the different occlusion conditions of different trips, the online local prior map contains more abundant supplementary information for the current driving scene. So we first search the historical prediction of different trips, and then search the historical prediction of the past time in the same trip as the local priors.

\begin{table}[t]\small
    \begin{center}
    \begin{threeparttable}
    \setlength{\tabcolsep}{1.3mm}
    \begin{tabular}{l|cccc}
    \toprule
     \textbf{Method}  &   $\textbf{AP}_{ped}$ & $\textbf{AP}_{div}$ & $\textbf{AP}_{bou}$ & \cellcolor{gray!30}\textbf{mAP}\\
    \midrule
    PriorDrive &  \textbf{71.8}  & \textbf{81.4} & \textbf{74.1} & \cellcolor{gray!30}\textbf{75.8}\\
    ~~~~w/o UVE  &   67.3 & 74.0 & 67.8 & \cellcolor{gray!30}69.7 \\
    ~~~~w/o Pretrain  &    69.4 & 76.5 & 69.6 & \cellcolor{gray!30}71.5\\ 
    ~~~~w/o SD Prior  &   69.5  & 77.1 & 72.2& \cellcolor{gray!30}72.9 \\
    ~~~~w/o Online Local  &  68.4  & 73.9 & 69.5 & \cellcolor{gray!30}70.6 \\
    ~~~~w/o HD Prior  &   67.3  & 72.0 & 68.6 & \cellcolor{gray!30}69.3 \\
    \bottomrule
    \end{tabular}
    \end{threeparttable}
    \end{center}

    \caption{The ablation of each component of PriorDrive.}
    \label{tab:abla}
\end{table}

\subsection{Main Results}

\begin{figure}[htbp]
\centering
\includegraphics[width=\columnwidth]{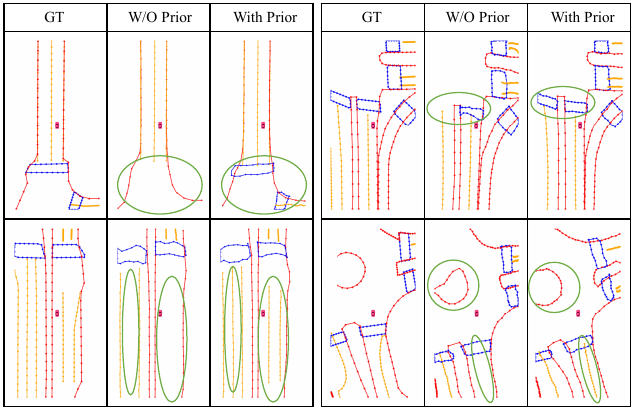}
\caption{Comparison of qualitative results between those with and without unified prior maps.
}
\label{fig:w/o_priors}

\end{figure}

\paragraph{Results on nuScenes.}
We evaluate our method across different model, evaluation metrics in \Cref{tab:main-result}. Our PriorDrive achieved a 7.2 mIoU improvement over baseline HDMapNet, which demonstrates that our vector prior remains effective even for rasterized models and metrics. In single-variable comparisons with prior/temporal-based methods employing MapTR as baseline, PriorDrive has demonstrated the most substantial improvement of 16 mAP. We further benchmarked our approach against the mainstream MapTRv2 baseline, which demonstrated an average performance gain greater than those achieved on HDMapNet, demonstrating that our HPQuery achieves more comprehensive and fine-grained exploitation of prior information. Furthermore, PriorDrive outperforms existing prior/temporal-based SOTA methods that adopt the MapTRv2 baseline, demonstrating the effectiveness of PriorDrive in fully utilizing complementary information from different prior maps. Figure~\ref{fig:w/o_priors} shows that using the prior maps can effectively recover the occluded elements, as well as more accurate mapping results. Overall, our method demonstrated consistent performance improvements across all baselines, highlighting its generalizability and potential applicability to other frameworks.

\paragraph{Results on Argoverse 2.}
In \Cref{tab:av2}, we present the results of 3 dim maps on Argoverse 2. PriorDrive improves performance by 8.1 mAP and outperforms the previous prior/temporal-based SOTA methods, demonstrating excellent generalization performance.

\paragraph{Results on OpenLane-V2.}
We compared PriorDrive on OpenLane-V2 when using ResNet-50 for training 24 epochs in \Cref{tab:openlanev2}. Compared to TopoLogic baseline, our method achieves better topological reasoning accuracy. And our overall score OLS reached 46.2 (v1.0.0) and 48.5 (v2.1.0), outperforming both Topologic and previous prior-based approaches. Overall, PriorDrive is a plug-and-play solution that can be widely applied to other vector tasks.

\subsection{Ablation Study}
we conducted ablation experiments on nuScenes based on MapTRv2 unless otherwise specified. More experiments are included in supplement.

\paragraph{Ablation Study of PriorDrive.} The ablation study in~\Cref{tab:abla} investigates the contribution of each component. Specifically, replacing the UVE with separate MLPs for each prior degrades performance by 6.1 mAP, demonstrating UVE's effectiveness in mitigating the inconsistency and weak alignment among different priors. indicating it can also alleviate these issues and enhance generalization. The performance degradation observed when ablating any single prior highlights that they contain complementary information, which PriorDrive is the first to leverage simultaneously.

\begin{table}[t]\small
    \begin{center}
    \begin{threeparttable}
    \setlength{\tabcolsep}{1.3mm}
    \begin{tabular}{l|cccc}
    \toprule
     \textbf{Method}  &   $\textbf{AP}_{ped}$ & $\textbf{AP}_{div}$ & $\textbf{AP}_{bou}$ & \cellcolor{gray!30}\textbf{mAP}\\
    \midrule
    StreamMapNet  &   29.6  & 30.1 & 41.9 & \cellcolor{gray!30}33.9 \\
    HRMapNet  &  36.9 & 30.3 & 44.0 & \cellcolor{gray!30}37.1 \\
    MapTracker  &  \textbf{45.9} & 30.0 & 45.1 & \cellcolor{gray!30}40.3 \\
    PriorDrive (Ours) &  44.7  & \textbf{35.5} & \textbf{47.3} & \cellcolor{gray!30}\textbf{42.5}\\
    \bottomrule
    \end{tabular}
    \end{threeparttable}
    \end{center}

    \caption{Comparisons on the new nuScenes split data sets proposed in StreamMapNet.}
    \label{tab:new_split}
\end{table}
\paragraph{Comparison on New Split.} The original split of nuScenes exhibits geographical overlap between training and validation sets. Therefore, we adopt new split proposed by StreamMapNet in~\Cref{tab:new_split}, which training and validation data are separated in locations. PriorDrive achieves 42.5 mAP, which outperforms the previous SOTA methods.

\begin{table}[t]\small
    \begin{center}
    \begin{threeparttable}
    \setlength{\tabcolsep}{0.4mm}
    \begin{tabular}{l|c|cccc}
    \toprule
     \textbf{Method}  &\textbf{Range}&   $\textbf{AP}_{ped}$ & $\textbf{AP}_{div}$ & $\textbf{AP}_{bou}$ & \cellcolor{gray!30}\textbf{mAP}\\
    \midrule
    P-MapNet&\multirow{4}{*}{120$\times$60m}&22.0&27.2&19.5&\cellcolor{gray!30}22.9\\
    StreamMapNet  &  &37.2  & 42.3 & 30.2 & \cellcolor{gray!30}36.6 \\
    HRMapNet  &  &43.1  & 47.7 & 34.9 & \cellcolor{gray!30}41.9 \\
    PriorDrive (Ours) & & \textbf{45.0}  & \textbf{49.5} & \textbf{37.3} & \cellcolor{gray!30}\textbf{43.9}\\
    \midrule
    P-MapNet&\multirow{4}{*}{240$\times$120m}&16.3&22.7&10.5&\cellcolor{gray!30}16.5\\
    StreamMapNet  &  &22.4  & 31.3 & 16.0 & \cellcolor{gray!30}23.3 \\
    HRMapNet  &  &\textbf{29.4}  & 34.2 & 19.7 & \cellcolor{gray!30}27.8 \\
    PriorDrive (Ours) & &28.8 &  \textbf{36.3} & \textbf{21.6} & \cellcolor{gray!30}\textbf{28.9}  \\
    
    \bottomrule
    \end{tabular}
    \end{threeparttable}
    \end{center}

    \caption{Performance comparison of large perception ranges on nuScenes.}
    \label{tab:large_range}
\end{table}

\paragraph{Comparison of Large Perception Ranges.}
Existing online mapping methods are constrained in prediction range due to the low resolution of distant areas in the captured images which leads to a significant decline in performance. However, PriorDrive leverages various prior maps to enhance accuracy in long-range predictions in~\Cref{tab:large_range}, demonstrating an outstanding foresight capability.

\section{Conclusion}
\label{sec:conclusion}
In this paper, we introduced PriorDrive, a novel framework that effectively and uniformly leverages various types of prior maps to enhance the accuracy and robustness of online HD map construction. Through comprehensive experiments, we demonstrated that UVE, combined with proposed pre-training strategies, significantly improves performance of state-of-the-art models in various vector tasks.
PriorDrive not only addresses the challenges associated with online HD map construction in complex environments but also offers a scalable solution that continually improves accuracy over time. The iterative use of historical prediction maps as priors leads to progressively refined map outputs, providing latest road information for autonomous driving.

\newpage
\bibliography{aaai2026}

@String(IJCV = {Int. J. Comput. Vis.})

@String(CVPR= {IEEE Conf. Comput. Vis. Pattern Recog.})

@String(ICCV= {Int. Conf. Comput. Vis.})

@String(ECCV= {Eur. Conf. Comput. Vis.})

@String(ICLR = {Int. Conf. Learn. Represent.})

@String(AAAI = {AAAI})

@String(IJCV  = {IJCV})

@String(CVPR  = {CVPR})

@String(ICCV  = {ICCV})

@String(ECCV  = {ECCV})

@String(ICLR  = {ICLR})

@ARTICLE{HDmaps-survey,
  author={Elghazaly, Gamal and Frank, Raphaël and Harvey, Scott and Safko, Stefan},
  journal={IEEE OJ-ITS}, 
  title={High-Definition Maps: Comprehensive Survey, Challenges, and Future Perspectives}, 
  year={2023},
  }

@INPROCEEDINGS{Gao2020VectorNetEH,
  title={VectorNet: Encoding HD Maps and Agent Dynamics From Vectorized Representation},
  author={Jiyang Gao and Chen Sun and Hang Zhao and Yi Shen and Dragomir Anguelov and Congcong Li and Cordelia Schmid},
  booktitle={CVPR},
  year={2020},
}

@inproceedings{Li2021HDMapNetAO,
  title={HDMapNet: An Online HD Map Construction and Evaluation Framework},
  author={Qi Li and Yue Wang and Yilun Wang and Hang Zhao},
  booktitle={ICRA},
  year={2021},
}

@inproceedings{Liu2022VectorMapNetEV,
  title={VectorMapNet: End-to-end Vectorized HD Map Learning},
  author={Yicheng Liu and Yue Wang and Yilun Wang and Hang Zhao},
  booktitle={ICML},
  year={2022},
}

@article{Jiang2024PMapNetFM,
  author       = {Zhou Jiang and
                  Zhenxin Zhu and
                  Pengfei Li and
                  Huan{-}ang Gao and
                  Tianyuan Yuan and
                  Yongliang Shi and
                  Hang Zhao and
                  Hao Zhao},
  title        = {P-MapNet: Far-seeing Map Generator Enhanced by both SDMap and HDMap Priors},
  journal={IEEE Robotics and Automation Letters}, 
  year         = {2024},
}

@InProceedings{schmidt2023navmap,
  author={Julian Schmidt and Julian Jordan and Franz Gritschneder and Thomas Monninger and Klaus Dietmayer},
  booktitle={ICRA}, 
  title={Exploring Navigation Maps for Learning-Based Motion Prediction}, 
  year={2023},
}

@article{Haklay2008OpenStreetMapUS,
  title={OpenStreetMap: User-Generated Street Maps},
  author={Mordechai (Muki) Haklay and Patrick Weber},
  journal={IEEE Pervasive Computing},
  year={2008},
}

@InProceedings{bevformer,
author="Li, Zhiqi
and Wang, Wenhai
and Li, Hongyang
and Xie, Enze
and Sima, Chonghao
and Lu, Tong
and Qiao, Yu
and Dai, Jifeng",
title="BEVFormer: Learning Bird's-Eye-View Representation from Multi-camera Images via Spatiotemporal Transformers",
booktitle="ECCV",
year="2022",
}

@inproceedings{detr,
  title={End-to-end object detection with transformers},
  author={Carion, Nicolas and Massa, Francisco and Synnaeve, Gabriel and Usunier, Nicolas and Kirillov, Alexander and Zagoruyko, Sergey},
  booktitle={ECCV},
  year={2020},
}

@InProceedings{Yuan_2024_streammapnet,
    author    = {Yuan, Tianyuan and Liu, Yicheng and Wang, Yue and Wang, Yilun and Zhao, Hang},
    title     = {StreamMapNet: Streaming Mapping Network for Vectorized Online HD Map Construction},
    booktitle = {WACV},
    year      = {2024},
}

@InProceedings{HIMAP,
    author    = {Zhou, Yi and Zhang, Hui and Yu, Jiaqian and Yang, Yifan and Jung, Sangil and Park, Seung-In and Yoo, ByungIn},
    title     = {HIMap: HybrId Representation Learning for End-to-end Vectorized HD Map Construction},
    booktitle = {CVPR},
    year      = {2024},
}

@InProceedings{Xiong2023NeuralMP,
  title={Neural Map Prior for Autonomous Driving},
  author={X. Xiong and Yicheng Liu and Tianyuan Yuan and Yue Wang and Yilun Wang and Hang Zhao},
  booktitle={CVPR},
  year={2023},
}

@article{MAPEX,
  title={Mind the map! Accounting for existing map information when estimating online HDMaps from sensor data},
  author={R'emy Sun and Li Yang and Diane Lingrand and Fr'ed'eric Precioso},
  journal={arXiv preprint arXiv:2311.10517},
  year={2023},
}

@inproceedings{betr,
  title={BERT: Pre-training of Deep Bidirectional Transformers for Language Understanding},
  author={Jacob Devlin and Ming-Wei Chang and Kenton Lee and Kristina Toutanova},
  booktitle={NAACL},
  year={2019},
}

@InProceedings{MAE,
  title={Masked Autoencoders Are Scalable Vision Learners},
  author={Kaiming He and Xinlei Chen and Saining Xie and Yanghao Li and Piotr Doll'ar and Ross B. Girshick},
  booktitle={CVPR},
  year={2021},
}

@InProceedings{liao2022maptr,
  title={MapTR: Structured Modeling and Learning for Online Vectorized HD Map Construction},
  author={Liao, Bencheng and Chen, Shaoyu and Wang, Xinggang and Cheng, Tianheng and Zhang, Qian and Liu, Wenyu and Huang, Chang},
  booktitle={ICLR},
  year={2023}
}

@InProceedings{Caesar2019nuScenesAM,
  title={nuScenes: A Multimodal Dataset for Autonomous Driving},
  author={Holger Caesar and Varun Bankiti and Alex H. Lang and Sourabh Vora and Venice Erin Liong and Qiang Xu and Anush Krishnan and Yuxin Pan and Giancarlo Baldan and Oscar Beijbom},
 booktitle={CVPR},
  year={2019},
}

@article{peng2024prevpredmap,
  title={PrevPredMap: Exploring Temporal Modeling with Previous Predictions for Online Vectorized HD Map Construction},
  author={Peng, Nan and Zhou, Xun and Wang, Mingming and Yang, Xiaojun and Chen, Songming and Chen, Guisong},
  journal={WACV},
  year={2025}
}

@InProceedings{BEMAPNET,
  title={End-to-End Vectorized HD-map Construction with Piecewise B{\'e}zier Curve},
  author={Limeng Qiao and Wenjie Ding and Xi Qiu and Chi Zhang},
  booktitle={CVPR},
  year={2023},
}

@InProceedings{PivotNet,
    author    = {Ding, Wenjie and Qiao, Limeng and Qiu, Xi and Zhang, Chi},
    title     = {PivotNet: Vectorized Pivot Learning for End-to-end HD Map Construction},
    booktitle = {ICCV},
    year      = {2023},
}

@inproceedings{MAPVR,
  title={Online map vectorization for autonomous driving: A rasterization perspective},
  author={Zhang, Gongjie and Lin, Jiahao and Wu, Shuang and Luo, Zhipeng and Xue, Yang and Lu, Shijian and Wang, Zuoguan and others},
  booktitle={NeurIPS},
  year={2024}
}

@article{Liu2019RoBERTaAR,
  title={RoBERTa: A Robustly Optimized BERT Pretraining Approach},
  author={Yinhan Liu and Myle Ott and Naman Goyal and Jingfei Du and Mandar Joshi and Danqi Chen and Omer Levy and Mike Lewis and Luke Zettlemoyer and Veselin Stoyanov},
  journal={arXiv preprint arXiv:1907.11692},
  year={2019},
}

@article{Lan2019ALBERTAL,
  title={ALBERT: A Lite BERT for Self-supervised Learning of Language Representations},
  author={Zhenzhong Lan and Mingda Chen and Sebastian Goodman and Kevin Gimpel and Piyush Sharma and Radu Soricut},
  journal={arXiv preprint arXiv:1909.11942},
  year={2019},
}

@inproceedings{Liu2022MixMAEMA,
  title={MixMAE: Mixed and Masked Autoencoder for Efficient Pretraining of Hierarchical Vision Transformers},
  author={Jihao Liu and Xin Huang and Jinliang Zheng and Yu Liu and Hongsheng Li},
  booktitle={CVPR},
  year={2023},

}

@article{maptrv2,
  title={MapTRv2: An End-to-End Framework for Online Vectorized HD Map Construction},
  author={Liao, Bencheng and Chen, Shaoyu and Zhang, Yunchi and Jiang, Bo and Zhang, Qian and Liu, Wenyu and Huang, Chang and Wang, Xinggang},
  journal={IJCV},
  year={2023}
}

@article{zeng2024driving,
  title={Driving with Prior Maps: Unified Vector Prior Encoding for Autonomous Vehicle Mapping},
  author={Zeng, Shuang and Chang, Xinyuan and Liu, Xinran and Pan, Zheng and Wei, Xing},
  journal={arXiv preprint arXiv:2409.05352},
  year={2024}
}

@inproceedings{TancikSMFRSRBN20,
  author       = {Matthew Tancik and
                  Pratul P. Srinivasan and
                  Ben Mildenhall and
                  Sara Fridovich{-}Keil and
                  Nithin Raghavan and
                  Utkarsh Singhal and
                  Ravi Ramamoorthi and
                  Jonathan T. Barron and
                  Ren Ng},
  title        = {Fourier Features Let Networks Learn High Frequency Functions in Low
                  Dimensional Domains},
  booktitle    = {NeurIPS},
  year         = {2020},
}

@article{fu2024topologic,
      title={TopoLogic: An Interpretable Pipeline for Lane Topology Reasoning on Driving Scenes}, 
      author={Yanping Fu and Wenbin Liao and Xinyuan Liu and Hang xu and Yike Ma and Feng Dai and Yucheng Zhang},
      year={2024},
      journal = {NeurIPS},
}

@article{wu2023topomlp,
  title={TopoMLP: An Simple yet Strong Pipeline for Driving Topology Reasoning},
  author={Wu, Dongming and Chang, Jiahao and Jia, Fan and Liu, Yingfei and Wang, Tiancai and Shen, Jianbing},
  journal={ICLR},
  year={2024}
}

@article{Liu2022PETRPE,
  title={PETR: Position Embedding Transformation for Multi-View 3D Object Detection},
  author={Yingfei Liu and Tiancai Wang and X. Zhang and Jian Sun},
  journal={ECCV},
  year={2022},
}

@inproceedings{zhang2024hrmapnet,
  title={Enhancing Vectorized Map Perception with Historical Rasterized Maps},
  author={Zhang, Xiaoyu and Liu, Guangwei and Liu, Zihao and Xu, Ningyi and Liu, Yunhui and Zhao, Ji},
  booktitle={ECCV},
  year={2024}
}

@article{zeng2025FSDrive,
      title={FutureSightDrive: Thinking Visually with Spatio-Temporal CoT for Autonomous Driving},
      author={Shuang Zeng and Xinyuan Chang and Mengwei Xie and Xinran Liu and Yifan Bai and Zheng Pan and Mu Xu and Xing Wei},
      journal={arXiv preprint arXiv:2505.17685},
      year={2025}
      }

@article{li2023toponet,
  title={Graph-based Topology Reasoning for Driving Scenes},
  author={Li, Tianyu and Chen, Li and Wang, Huijie and Li, Yang and Yang, Jiazhi and Geng, Xiangwei and Jiang, Shengyin and Wang, Yuting and Xu, Hang and Xu, Chunjing and Yan, Junchi and Luo, Ping and Li, Hongyang},
  journal={arXiv preprint arXiv:2304.05277},
  year={2023}
}

@INPROCEEDINGS { Argoverse2,
  author = {Benjamin Wilson and William Qi and Tanmay Agarwal and John Lambert and Jagjeet Singh and Siddhesh Khandelwal and Bowen Pan and Ratnesh Kumar and Andrew Hartnett and Jhony Kaesemodel Pontes and Deva Ramanan and Peter Carr and James Hays},
  title = {Argoverse 2: Next Generation Datasets for Self-Driving Perception and Forecasting},
  booktitle = {NeurIPS},
  year = {2021}
}

@inproceedings{wang2023openlanev2,
  title={OpenLane-V2: A Topology Reasoning Benchmark for Unified 3D HD Mapping}, 
  author={Wang, Huijie and Li, Tianyu and Li, Yang and Chen, Li and Sima, Chonghao and Liu, Zhenbo and Wang, Bangjun and Jia, Peijin and Wang, Yuting and Jiang, Shengyin and Wen, Feng and Xu, Hang and Luo, Ping and Yan, Junchi and Zhang, Wei and Li, Hongyang},
  booktitle={NeurIPS},
  year={2023}
}

@article{osmr,
  title={Enhancing Online Road Network Perception and Reasoning with Standard Definition Maps},
  author={Hengyuan Zhang and David Paz and Yuliang Guo and Arun Das and Xinyu Huang and Karsten Haug and Henrik I. Christensen and Liu Ren},
  journal={IROS},
  year={2024},
}

@article{Ma2024RoadPainterPA,
  title={RoadPainter: Points Are Ideal Navigators for Topology transformER},
  author={Zhongxing Ma and Shuang Liang and Yongkun Wen and Weixin Lu and Guowei Wan},
  journal={ECCV},
  year={2024}
}

@article{SMERF,
  title={Augmenting Lane Perception and Topology Understanding with Standard Definition Navigation Maps},
  author={Katie Z Luo and Xinshuo Weng and Yan Wang and Shuang Wu and Jie Li and Kilian Q. Weinberger and Yue Wang and Marco Pavone},
  journal={ICRA},
  year={2024}
}

@inproceedings{Shelhamer2014FullyCN,
  title={Fully convolutional networks for semantic segmentation},
  author={Evan Shelhamer and Jonathan Long and Trevor Darrell},
  booktitle={CVPR},
  year={2015},

}

@article{zhang2024self,
  title={Self-Adaptive Robust Motion Planning for High DoF Robot Manipulator using Deep MPC},
  author={Zhang, Ye and Mo, Kangtong and Shen, Fangzhou and Xu, Xuanzhen and Zhang, Xingyu and Yu, Jiayue and Yu, Chang},
  journal={arXiv preprint arXiv:2407.12887},
  year={2024}
}

@ARTICLE{yao2024waterscenes,
  author={Yao, Shanliang and Guan, Runwei and Wu, Zhaodong and Ni, Yi and Huang, Zile and Liu, Ryan Wen and Yue, Yong and Ding, Weiping and Lim, Eng Gee and Seo, Hyungjoon and Man, Ka Lok and Ma, Jieming and Zhu, Xiaohui and Yue, Yutao},
  journal={IEEE Transactions on Intelligent Transportation Systems}, 
  title={WaterScenes: A Multi-Task 4D Radar-Camera Fusion Dataset and Benchmarks for Autonomous Driving on Water Surfaces}, 
  year={2024},
}

@article{yao2025exploring,
  title={Exploring radar data representations in autonomous driving: A comprehensive review},
  author={Yao, Shanliang and Guan, Runwei and Peng, Zitian and Xu, Chenhang and Shi, Yilu and Ding, Weiping and Lim, Eng Gee and Yue, Yong and Seo, Hyungjoon and Man, Ka Lok and others},
  journal={IEEE Transactions on Intelligent Transportation Systems},
  year={2025},
}

@INPROCEEDINGS{yao2025usvtrack,
      title={USVTrack: USV-Based 4D Radar-Camera Tracking Dataset for Autonomous Driving in Inland Waterways},
      booktitle={IROS}, 
      author={Shanliang Yao and Runwei Guan and Yi Ni and Sen Xu and Yong Yue and Xiaohui Zhu and Ryan Wen Liu},
      year={2025}
}

@article{sun2023large,
  title={Large language models as topological structure enhancers for text-attributed graphs},
  author={Sun, Shengyin and Ren, Yuxiang and Ma, Chen and Zhang, Xuecang},
  journal={arXiv preprint arXiv:2311.14324},
  year={2023}
}

@ARTICLE{9681188,
  author={Huang, Zhenhua and Yang, Shunzhi and Zhou, MengChu and Li, Zhetao and Gong, Zheng and Chen, Yunwen},
  journal={IEEE Transactions on Image Processing}, 
  title={Feature Map Distillation of Thin Nets for Low-Resolution Object Recognition}, 
  year={2022},
  }

@article{hu2024tmff,
  title={TMFF: Trustworthy Multi-Focus Fusion Framework for Multi-Label Sewer Defect Classification in Sewer Inspection Videos},
  author={Hu, Chuanfei and Zhao, Chenyang and Shao, Hang and Deng, Jin and Wang, Yongxiong},
  journal={IEEE Transactions on Circuits and Systems for Video Technology},
  year={2024},
}

@article{wu2025interactionmap,
  title={InteractionMap: Improving Online Vectorized HDMap Construction with Interaction},
  author={Wu, Kuang and Yang, Chuan and Li, Zhanbin},
  journal={CVPR},
  year={2025}
}

@article{202506.2087,
	year = 2025,
	author = {Zhichao Ma and Yutong Luo and Zheyu Zhang and Aijia Sun and Yinuo Yang and Hao Liu},
	title = {Reinforcement Learning Approach for Highway Lane-Changing: PPO-Based Strategy Design},
	journal = {Preprints}
}

@article{li2024towards1,
  title={Towards Visual-Prompt Temporal Answer Grounding in Instructional Video},
  author={Li, Shutao and Li, Bin and Sun, Bin and Weng, Yixuan},
  journal={IEEE transactions on pattern analysis and machine intelligence},
  volume={46},
  number={12},
  pages={8836--8853},
  year={2024}
}

@article{xie2025seqgrowgraph,
  title={SeqGrowGraph: Learning Lane Topology as a Chain of Graph Expansions},
  author={Xie, Mengwei and Zeng, Shuang and Chang, Xinyuan and Liu, Xinran and Pan, Zheng and Xu, Mu and Wei, Xing},
  journal={ICCV},
  year={2025}
}

@article{lu2025differentiable,
  title={Differentiable NMS via Sinkhorn Matching for End-to-End Fabric Defect Detection},
  author={Lu, Zhengyang and Lu, Bingjie and Wang, Weifan and Wang, Feng},
  journal={arXiv preprint arXiv:2505.07040},
  year={2025}
}

@article{guo2025depth,
  title={Depth-Aware Super-Resolution via Distance-Adaptive Variational Formulation},
  author={Guo, Tianhao and Lu, Bingjie and Wang, Feng and Lu, Zhengyang},
  journal={arXiv preprint arXiv:2509.05746},
  year={2025}
}

@article{zeng2025janusvln,
            title={JanusVLN: Decoupling Semantics and Spatiality with Dual Implicit Memory for Vision-Language Navigation},
            author={Zeng, Shuang and Qi, Dekang and Chang, Xinyuan and Xiong, Feng and Xie, Shichao and Wu, Xiaolong and Liang, Shiyi and Xu, Mu and Wei, Xing},
            journal={arXiv preprint arXiv:2509.22548},
            year={2025}
            }

@article{xiao2025worldenv,
      title={World-Env: Leveraging World Model as a Virtual Environment for VLA Post-Training}, 
      author={Junjin Xiao and Yandan Yang and Xinyuan Chang and Ronghan Chen and Feng Xiong and Mu Xu and Wei-Shi Zheng and Qing Zhang},
      year={2025},
      journal={arXiv preprint arXiv:2509.24948},
}

@article{wan2025driving,
  title={Driving by Hybrid Navigation: An Online HD-SD Map Association Framework and Benchmark for Autonomous Vehicles},
  author={Wan, Jiaxu and Wang, Xu and Xie, Mengwei and Chang, Xinyuan and Liu, Xinran and Pan, Zheng and Xu, Mu and Yuan, Ding},
  journal={arXiv preprint arXiv:2507.07487},
  year={2025}
}

@article{yuan2025unimapgen,
  title={UniMapGen: A Generative Framework for Large-Scale Map Construction from Multi-modal Data},
  author={Yuan, Yujian and Wu, Changjie and Chang, Xinyuan and Wang, Sijin and Zhang, Hang and Liang, Shiyi and Zeng, Shuang and Xu, Mu},
  journal={arXiv preprint arXiv:2509.22262},
  year={2025}
}

@article{liang2025persistent,
  title={Persistent Autoregressive Mapping with Traffic Rules for Autonomous Driving},
  author={Liang, Shiyi and Chang, Xinyuan and Wu, Changjie and Yan, Huiyuan and Bai, Yifan and Liu, Xinran and Zhang, Hang and Yuan, Yujian and Zeng, Shuang and Xu, Mu and others},
  journal={arXiv preprint arXiv:2509.22756},
  year={2025}
}

@inproceedings{tang2025ocrt,
  title={OCRT: Boosting Foundation Models in the Open World with Object-Concept-Relation Triad},
  author={Tang, Luyao and Yuan, Yuxuan and Chen, Chaoqi and Zhang, Zeyu and Huang, Yue and Zhang, Kun},
  booktitle={Proceedings of the Computer Vision and Pattern Recognition Conference},
  year={2025}
}

@article{tang2025dissecting,
  title={Dissecting Generalized Category Discovery: Multiplex Consensus under Self-Deconstruction},
  author={Tang, Luyao and Huang, Kunze and Chen, Chaoqi and Yuan, Yuxuan and Li, Chenxin and Tu, Xiaotong and Ding, Xinghao and Huang, Yue},
  journal={arXiv preprint arXiv:2508.10731},
  year={2025}
}

@inproceedings{huang2024ar,
  title={AR Overlay: Training Image Pose Estimation on Curved Surface in a Synthetic Way},
  author={Huang, Sining and Song, Yukun and Kang, Yixiao and Yu, Chang and others},
  booktitle={CS \& IT Conference Proceedings},
  year={2024},
}

@article{202508.0462,
    year = 2025,
    author = {Sining Huang and Geyu Shen and Yixiao Kang and Yukun Song},
    title = {Immersive Augmented Reality Music Interaction through Spatial Scene Understanding and Hand Gesture Recognition},
    journal = {Preprints}
}

@Article{rs17132267,
AUTHOR = {Al Shafian, Sultan and He, Chao and Hu, Da},
TITLE = {DamageScope: An Integrated Pipeline for Building Damage Segmentation, Geospatial Mapping, and Interactive Web-Based Visualization},
JOURNAL = {Remote Sensing},
YEAR = {2025},
}

@article{he2025social,
  title={Social Media Analytics for Disaster Response: Classification and Geospatial Visualization Framework},
  author={He, Chao and Hu, Da},
  journal={Applied Sciences},
  year={2025},
}

@article{ren2025digital,
  title={Digital Genealogy: AIGC-driven Evolution of Digital Twin for Future Smart Manufacturing},
  author={Ren, Lei and Dong, Jiabao and Zeng, Xianchao and Yang, Lingyuan and Wang, Yuqing},
  journal={IEEE Transactions on Automation Science and Engineering},
  year={2025},

}

@misc{zeng2025lens,
      title={LENSLLM: Unveiling Fine-Tuning Dynamics for LLM Selection}, 
      author={Xinyue Zeng and Haohui Wang and Junhong Lin and Jun Wu and Tyler Cody and Dawei Zhou},
      year={2025},
journal={arXiv preprint arXiv:2505.03793},

}

@misc{lin2025plan,
      title={Plan and Budget: Effective and Efficient Test-Time Scaling on Large Language Model Reasoning}, 
      author={Junhong Lin and Xinyue Zeng and Jie Zhu and Song Wang and Julian Shun and Jun Wu and Dawei Zhou},
      year={2025},
journal={arXiv preprint arXiv:2505.16122},
}

@article{yang2025unleashing,
  title={Unleashing the potential of multimodal llms for zero-shot spatio-temporal video grounding},
  author={Yang, Zaiquan and Liu, Yuhao and Hancke, Gerhard and Lau, Rynson WH},
  journal={arXiv preprint arXiv:2509.15178},
  year={2025}
}

@inproceedings{zhou2025hademif,
  title={Hademif: Hallucination detection and mitigation in large language models},
  author={Zhou, Xiaoling and Zhang, Mingjie and Lee, Zhemg and Ye, Wei and Zhang, Shikun},
  booktitle={ICLR},
  year={2025}
}

@article{zhou2025valuing,
  title={Valuing training data via causal inference for in-context learning},
  author={Zhou, Xiaoling and Ye, Wei and Lee, Zhemg and Zou, Lei and Zhang, Shikun},
  journal={IEEE Transactions on Knowledge and Data Engineering},
  year={2025},
}

@inproceedings{li2025revolutionizing,
  title={Revolutionizing Drug Discovery: Integrating Spatial Transcriptomics with Advanced Computer Vision Techniques},
  author={Li, Zichao and Qiu, Shiqing and Ke, Zong},
  booktitle={1st CVPR Workshop on Computer Vision For Drug Discovery (CVDD): Where are we and What is Beyond?},
  year={2025}
}

@inproceedings{li2025knowledge,
  title={Knowledge-Grounded Detection of Cryptocurrency Scams with Retrieval-Augmented LMs},
  author={Li, Zichao},
  booktitle={Knowledgeable Foundation Models at ACL 2025},
  year={2025},
}

@article{chen2025visrl,
  title={Visrl: Intention-driven visual perception via reinforced reasoning},
  author={Chen, Zhangquan and Luo, Xufang and Li, Dongsheng},
  journal={arXiv preprint arXiv:2503.07523},
  year={2025}
}

@article{chen2025sifthinker,
  title={SIFThinker: Spatially-Aware Image Focus for Visual Reasoning},
  author={Chen, Zhangquan and Zhao, Ruihui and Luo, Chuwei and Sun, Mingze and Yu, Xinlei and Kang, Yangyang and Huang, Ruqi},
  journal={arXiv preprint arXiv:2508.06259},
  year={2025}
}

@article{sun2025yolov4svm,
  title={A Lightweight YOLOv4-SVM Model for Automated Waste Monitoring in Smart Cities},
  author={Sun, Qianyi and Li, Jiaxuan},
  journal={TechRxiv preprint},
  year={2025},
}

@inproceedings{lan2025contextual,
title={Contextual Integrity in {LLM}s via Reasoning and Reinforcement Learning},
author={Lan, Guangchen and Inan, Huseyin A and Abdelnabi, Sahar and Kulkarni, Janardhan and Wutschitz, Lukas and Shokri, Reza and Brinton, Christopher G and Sim, Robert},
booktitle={NeurIPS},
year={2025}
}

@article{lan2025mappo,
  title={MaPPO: Maximum a Posteriori Preference Optimization with Prior Knowledge},
  author={Lan, Guangchen and Zhang, Sipeng and Wang, Tianle and Zhang, Yuwei and Zhang, Daoan and Wei, Xinpeng and Pan, Xiaoman and Zhang, Hongming and Han, Dong-Jun and Brinton, Christopher G},
  journal={arXiv preprint arXiv:2507.21183},
  year={2025}
}

@article{cui2025efficient,
  title={Efficient Localization and Spatial Distribution Modeling of Canopy Palms Using UAV Imagery},
  author={Cui, Kangning and Tang, Wei and Zhu, Rongkun and Wang, Manqi and Larsen, Gregory D and Pauca, Victor P and Alqahtani, Sarra and Yang, Fan and Segurado, David and Fine, Paul and others},
  journal={IEEE Transactions on Geoscience and Remote Sensing},
  year={2025},
}

@article{jin2025reasoning,
  title={Reasoning or Not? A Comprehensive Evaluation of Reasoning LLMs for Dialogue Summarization},
  author={Jin, Keyan and Wang, Yapeng and Santos, Leonel and Fang, Tao and Yang, Xu and Im, Sio Kei and Oliveira, Hugo Gon{\c{c}}alo},
  journal={arXiv preprint arXiv:2507.02145},
  year={2025}
}

@article{li2025multi,
  title={Multi-Objective Unlearning in Recommender Systems via Preference Guided Pareto Exploration},
  author={Li, Yuyuan and Zhang, Yizhao and Liu, Weiming and Feng, Xiaohua and Han, Zhongxuan and Chen, Chaochao and Yan, Chenggang},
  journal={IEEE Transactions on Services Computing},
  year={2025},
}

@article{li2023ultrare,
  title={Ultrare: Enhancing receraser for recommendation unlearning via error decomposition},
  author={Li, Yuyuan and Chen, Chaochao and Zhang, Yizhao and Liu, Weiming and Lyu, Lingjuan and Zheng, Xiaolin and Meng, Dan and Wang, Jun},
  journal={NeurIPS},
  year={2023}
}

@inproceedings{dai2025unbiased,
  title={Unbiased Missing-modality Multimodal Learning},
  author={Dai, Ruiting and Li, Chenxi and Yan, Yandong and Mo, Lisi and Qin, Ke and He, Tao},
  booktitle={ICCV},
  year={2025}
}

@inproceedings{yin2025knowledge,
  title={Knowledge-Aligned Counterfactual-Enhancement Diffusion Perception for Unsupervised Cross-Domain Visual Emotion Recognition},
  author={Yin, Wen and Wang, Yong and Duan, Guiduo and Zhang, Dongyang and Hu, Xin and Li, Yuan-Fang and He, Tao},
  booktitle={CVPR},
  year={2025}
}

@inproceedings{
zhang2025molebridge,
title={MoleBridge: Synthetic Space Projecting with Discrete Markov Bridges},
author={Rongchao Zhang and Yu Huang and Yongzhi Cao and Hanpin Wang},
booktitle={NeurIPS},
year={2025},

}

@article{ZHANG2025103888,
title = {StrFilter: Multi-Modal Medical Image Fusion via Structure-Oriented Adaptive Filtering},
journal = {Information Fusion},
year = {2025},
author = {Rongchao Zhang and Weiping Ding and Hongbin Han and Yongzhi Cao and Hanpin Wang and Yu Huang},

}

@inproceedings{wei2025copeft,
  title={CoPEFT: Fast Adaptation Framework for Multi-Agent Collaborative Perception with Parameter-Efficient Fine-Tuning},
  author={Wei, Quanmin and Dai, Penglin and Li, Wei and Liu, Bingyi and Wu, Xiao},
  booktitle={AAAI},
  year={2025}
}

@inproceedings{wei2026infocom,
  title={InfoCom: Kilobyte-Scale Communication-Efficient Collaborative Perception with Information Bottleneck},
  author={Wei, Quanmin and Dai, Penglin and Li, Wei and Liu, Bingyi and Wu, Xiao},
  booktitle={AAAI},
  year={2026}
}

@article{li2025cogvla,
  title={Cogvla: Cognition-aligned vision-language-action model via instruction-driven routing \& sparsification},
  author={Li, Wei and Zhang, Renshan and Shao, Rui and He, Jie and Nie, Liqiang},
  journal={arXiv preprint arXiv:2508.21046},
  year={2025}
}

@inproceedings{li2025lion,
  title={Lion-fs: Fast \& slow video-language thinker as online video assistant},
  author={Li, Wei and Hu, Bing and Shao, Rui and Shen, Leyang and Nie, Liqiang},
  booktitle={CVPR},
  year={2025}
}

@inproceedings{10.1117/12.3049486,
author = {Yixiao Yuan and Yangchen Huang and Yu Ma and Xinjin Li and Zhenglin Li and Yiming Shi and Huapeng Zhou},
title = {{Rhyme-aware Chinese lyric generator based on GPT}},
booktitle = {AANN},
year = {2024},

}

@misc{li2025hyfacialhybrid,
title={Hy-Facial: Hybrid Feature Extraction by Dimensionality Reduction Methods for Enhanced Facial Expression Classification},
author={Xinjin Li and Yu Ma and Kaisen Ye and Jinghan Cao and Minghao Zhou and Yeyang Zhou},
year={2025},
  journal={arXiv preprint arXiv:2509.26614},

}

\newpage
\appendix

\twocolumn[
\begin{center}
  {\LARGE \bfseries Supplementary Material \par}
  \vspace{1em}
\end{center}
]

\section{More Experiments}
Here, we present a detailed examination of additional ablation studies on MapTRv2 unless otherwise specified.

\begin{table}[t]
    \begin{center}
    \begin{threeparttable}
    \setlength{\tabcolsep}{1.0mm}
    \begin{tabular}{ccc|cccc}
    \toprule
    \textbf{M}&  \textbf{N} & \multirow{2}{*}{\textbf{Dim}}&  \multicolumn{4}{c}{\textbf{AP}}  \\
     \textbf{Intra}&\textbf{Inter} &  &  \textbf{Ped.} & \textbf{Div.} & \textbf{Bou.} & \cellcolor{gray!30}\textbf{Mean}  \\
     \midrule
     \multicolumn{3}{c|}{w/o UVE}  & 67.3 & 74.0 & 67.8 & \cellcolor{gray!30}69.7 \\
     \midrule
   4 & 4 & 64 & 71.3 & 80.0 & \textbf{74.2} & \cellcolor{gray!30}75.2 \\
   2 & 2 & 128& 71.0 & 80.2 & 73.5 & \cellcolor{gray!30}74.9  \\ 
   2 & 2 & 64 &  \textbf{71.8}& \textbf{81.4} & 74.1 & \cellcolor{gray!30}\textbf{75.8}  \\
   1 & 2 & 64 & 70.2 & 79.1 & 73.2 & \cellcolor{gray!30}74.2 \\
   2 & 1 & 64 & 70.4 & 78.5 & 71.9 & \cellcolor{gray!30}73.6 \\
   1 & 1 & 32 & 68.1 & 76.3 & 69.6 & \cellcolor{gray!30}71.3  \\
    \bottomrule
    \end{tabular}
    \end{threeparttable}
    \end{center}
    \vspace{-0.3cm}
    \caption{Ablation experiments of UVE structure.}
    \label{tab:arch_UVE}
    \vspace{-0.1cm}
\end{table}          

\paragraph{Structure of UVE.} \Cref{tab:arch_UVE} presents the results of ablation experiments on the UVE structure. As the UVE model parameters are progressively augmented, the mAP increases to 75.8, while the model maintains a lightweight architecture. However, performance saturates with further parameter increases, constrained by the limited volume and diversity of the vector data.

\begin{table}[t]\small
    \centering
    \begin{threeparttable}
    \setlength{\tabcolsep}{1.0mm}
    \begin{tabular}{ccc|cccc}
    \toprule
    \multicolumn{3}{c|}{\textbf{Method}}   &   \multicolumn{4}{c}{\textbf{AP}}  \\
    \textbf{Type} & \textbf{Seg} & \textbf{Pt} &    \textbf{Ped.} & \textbf{Div.} & \textbf{Bou.} & \cellcolor{gray!30}\textbf{Mean}  \\
    \midrule
    \multicolumn{3}{c|}{w/o Pretrain}  & 69.4 & 76.5 & 69.6 & \cellcolor{gray!30}71.5\\ 
    \midrule   
   Original & - & -  & 69.9 & 77.1 & 70.2 & \cellcolor{gray!30}72.4  \\
   Noise~3m & 20\% & -  & 69.8 & 78.0 & 71.7 & \cellcolor{gray!30}73.2 \\
   Noise~1m & 10\% & -  & 71.0 & 79.2 & 72.7 & \cellcolor{gray!30}74.3 \\
   Noise~1m & 10\% & 5\%  & \textbf{72.0} & 80.1 & 73.5 & \cellcolor{gray!30}75.2 \\   
   Mask & 10\% & 5\%  & 71.4 & 79.8 & 73.5 & \cellcolor{gray!30}74.9  \\   
   Mask \& Noise~1m & 10\% & 5\%  & 71.8 & \textbf{81.4} & \textbf{74.1} &  \cellcolor{gray!30}\textbf{75.8} \\ 
   
    \bottomrule
    \end{tabular}
    \end{threeparttable}
    \caption{Ablation of the pre-training strategies. “Seg” and “Pt” represent segment-level and point-level, respectively.}
    \label{tab:pretrain}
\end{table}

\paragraph{The Effect of Pre-training.} To effectively capture the prior information of vector maps, we pre-trained UVE using different strategies in \Cref{tab:pretrain}. When we fed the original vector map directly to UVE without any noise, the simple task might have limited the model's ability to fully learn the prior distribution, resulting in only a 0.9 mAP improvement.
We randomly add segment-level or point-level Gaussian noise to map elements. Among them, 10\%  segment-level noise and 5\% point-level noise increase 3.7 mAP the most. This demonstrates that strategy incorporating both segment-level and point-level noise are more effective at learning the prior information.
We also experimented with masking instead of noise, which resulted in a 3.4 mAP improvement. Finally, we conducted pre-training by simultaneously applying gaussian noise and mask, achieving the best result of 75.8 mAP.

\begin{table}[t]\small
    \centering
    \begin{threeparttable}
    \setlength{\tabcolsep}{1.3mm}
    \begin{tabular}{l|c|cccc}
    \toprule
    \textbf{Fixed} &\textbf{Prior} &   \multicolumn{4}{c}{\textbf{AP}}  \\
    \textbf{Parameters}&\textbf{Num} &    \textbf{Ped.} & \textbf{Div.} & \textbf{Bou.} & \cellcolor{gray!30}\textbf{Mean}  \\
    \midrule
    w/o online local & None &  68.4  & 73.9 & 69.5 & \cellcolor{gray!30}70.6 \\
    \midrule
    \cellcolor[rgb]{ .906,  .902,  .902}\emph{search\_range=5m} & One &  70.6  & 78.6 & 71.7 & \cellcolor{gray!30}73.6 \\
    \cellcolor[rgb]{ .906,  .902,  .902}\emph{pretrain=true} & Two &  70.6  & 79.2 &72.4 & \cellcolor{gray!30}74.1 \\
    \cellcolor[rgb]{ .906,  .902,  .902}\emph{integration=replace} & Multiple &  \textbf{71.8}  & \textbf{81.4} & \textbf{74.1} & \cellcolor{gray!30}\textbf{75.8} \\
    \bottomrule
    \end{tabular}
    \end{threeparttable}
    \caption{Ablation study on the number of online local prior maps.}
    \label{tab:prior_num}
\end{table}

\paragraph{Impact of the Number of Prior Maps.}
In history, different vehicles passed through the same area. Due to the varying degrees of obstruction, these vehicles would predict different online local maps, thus being able to be used as prior maps. Also, the online local maps predicted by the same vehicle a few seconds ago also have reference value, so they can also be used as prior maps. Therefore, we conducted an ablation study to assess the effect of the number of online local prior maps (see \Cref{tab:prior_num}). The results indicate that using a single online local prior map led to a 3.0 mAP improvement, demonstrating its effectiveness in enhancing current perception. Incorporating two prior maps resulted in a more substantial improvement of 3.5 mAP, reflecting the value of richer prior information. Due to the varying number of available online local prior maps in different regions, and because of the hyperparameter settings of UVE that can encode vector points, the number of multiple online local prior maps is generally 3 to 5.
Multiple prior maps effectively improved 5.2 mAP, indicating that our UVE can adaptively alleviate potential inconsistencies across different online local maps and more prior maps can provide more comprehensive necessary information.

\begin{table}[t]
    \centering
    \begin{threeparttable}
    \setlength{\tabcolsep}{1.5mm}
    \begin{tabular}{l|c|cccc}
    \toprule
    \textbf{Fixed}&\textbf{Search}  &   \multicolumn{4}{c}{\textbf{AP}}  \\
    \textbf{Parameters} & \textbf{Range} &    \textbf{Ped.} & \textbf{Div.} & \textbf{Bou.} & \cellcolor{gray!30}\textbf{Mean}  \\
    \midrule
    w/o online local & None &  68.4  & 73.9 & 69.5 & \cellcolor{gray!30}70.6 \\
     \midrule 
    \cellcolor[rgb]{ .906,  .902,  .902}\emph{prior\_num=2}    &5 m & \textbf{71.8}  & \textbf{81.4} & \textbf{74.1} & \cellcolor{gray!30}\textbf{75.8} \\
    \cline{2-6} 
    \cellcolor[rgb]{ .906,  .902,  .902}\emph{pretrain=true} &10 m & 70.2 & 79.4 & 72.7 & \cellcolor{gray!30}74.1  \\
    \cline{2-6} 
    \cellcolor[rgb]{ .906,  .902,  .902}\emph{integration=replace} &15 m &69.0 & 76.7 & 70.8 & \cellcolor{gray!30}72.2  \\    
    \bottomrule
    \end{tabular}
    \end{threeparttable}
    \caption{Ablation experiments on the search range of online local prior maps on MapTRv2.}
    \label{tab:range}
\end{table}

\paragraph{Effect of different search ranges for prior maps.}
We search for available online local prior maps based on the distance between the geographical location when the online local prior map was generated and the current geographical location of the vehicle. So we performed ablation experiments to explore the impact of search ranges for prior maps in \Cref{tab:range}. The results show that as the search range is narrowed, the effect gradually improves. Although a smaller search range may reduce the number of online local prior maps, it will be more accurate in terms of geographical overlap. In such a situation of slight inconsistency and misalignment, our PriorDrive can effectively alleviate the problem.

\begin{table}[t]
    \centering
    \begin{threeparttable}
    \setlength{\tabcolsep}{1.2mm}
    \begin{tabular}{l|c|cccc}
    \toprule
   \textbf{Fixed} &\textbf{Integrated} &   \multicolumn{4}{c}{\textbf{AP}}  \\
    \textbf{Parameters} &\textbf{Method} &   \textbf{Ped.} & \textbf{Div.} & \textbf{Bou.} & \cellcolor{gray!30}\textbf{Mean}   \\
    \midrule
    w/o online local & None &  68.4  & 73.9 & 69.5 & \cellcolor{gray!30}70.6 \\
     \midrule    
    \cellcolor[rgb]{ .906,  .902,  .902}\emph{prior\_num=2} & Replace & 71.5 & 79.6 & 73.1 &  \cellcolor{gray!30}74.7 \\
    \cline{2-6} 
    \cellcolor[rgb]{ .906,  .902,  .902}\emph{pretrain=true} & Add & \textbf{71.8} & 80.2 & 73.6 &  \cellcolor{gray!30}75.2 \\
    \cline{2-6} 
    \cellcolor[rgb]{ .906,  .902,  .902}\emph{search\_range=5m} & Concat &\textbf{71.8}  & \textbf{81.4} & \textbf{74.1} & \cellcolor{gray!30}\textbf{75.8} \\
    \bottomrule
    \end{tabular}
    \end{threeparttable}
    \caption{Ablation experiments on the integrated methods of online local prior maps on MapTRv2.}
    \label{tab:integ}
\end{table}

\paragraph{Different methods for integrating prior maps.}
To evaluate the versatility of our approach, we tested various integration methods in \Cref{tab:integ}. Compared to MapTRv2, the HPQuery methods of replacement and addition resulted in mAP improvements of 4.1 and 4.6, respectively, but these methods may impact the performance of the original query. In contrast, the concatenation method achieved a more substantial improvement of 5.2 mAP. This approach effectively preserves the original query's integrity while also leveraging the instance-level and point-level prior features learned by UVE, making it the most effective integration method.

\begin{table}[t]\small
\centering
    \begin{threeparttable}
    \setlength{\tabcolsep}{0.5mm}
    \begin{tabular}{l|c|c}
    \toprule
    & \textbf{MapTRv2} & \textbf{Ours}     \\
     \midrule
    FPS & 10.3&9.9 \\
    GPU mem (MB) & 2624 & 2650\\
    Params (MB) & 40.3 & 43.4\\
    mAP  &61.5&75.8 \\
    \bottomrule
    \toprule
    \textbf{Component}&\textbf{Runtime (ms)} & \textbf{Proportion}     \\
     \midrule
    2D Backbone &47.6 & 50.2\%\\
    PV to BEV & 12.9 & 13.6\%\\
    UVE & 14.6 & 15.3\%\\
    Map Decoder & 19.8 &20.9\% \\
    \bottomrule
    \end{tabular}
    \end{threeparttable}
    \caption{Detailed comparison of our method with MapTRv2. }
    \label{tab:FPS}
\end{table}

\paragraph{Inference Speed, Memory, Model Size, and Runtime.} The results in \Cref{tab:FPS} measured on a single RTX A6000 show that our method maintains similar FPS, GPU memory usage, and parameter count compared to MapTRv2, while improving 14.3 mAP. Our UVE accounts for only 15.3\% of the total pipeline, highlighting the efficiency and practical value of our approach across various application scenarios.

\section{More Qualitative Visualization}

\begin{figure*}[!b]
\centering
\includegraphics[width=0.8\textwidth]{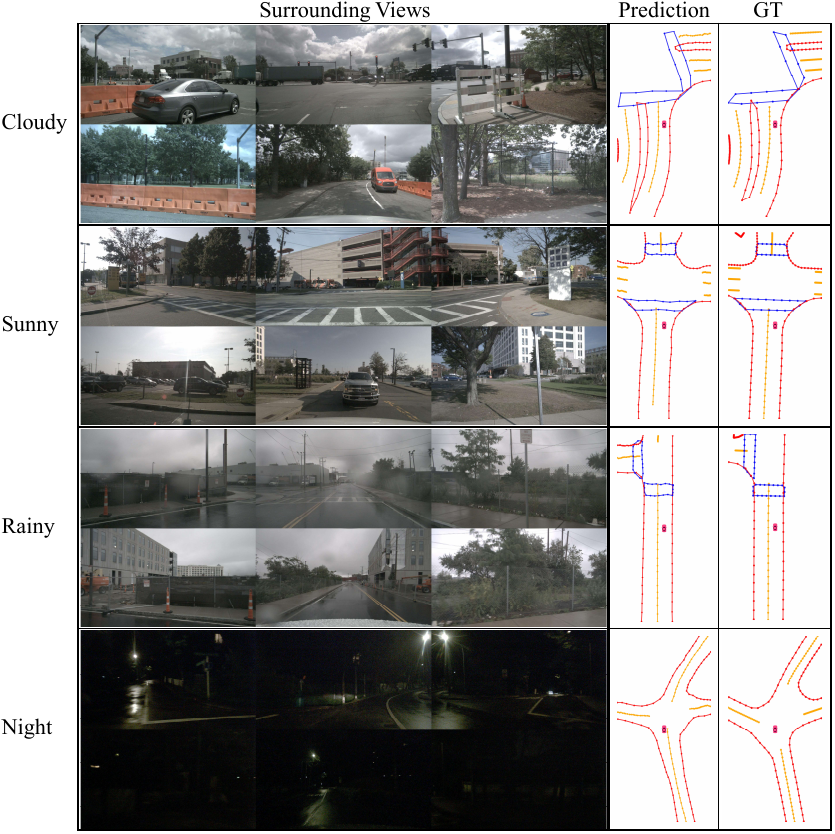}
\caption{Our method maintains stable and robust HD map construction quality in complex and variable driving scenarios on nuScenes.}
\label{fig:scenarios}
\end{figure*}

\begin{figure*}[h]
\centering
\includegraphics[width=0.8\textwidth]{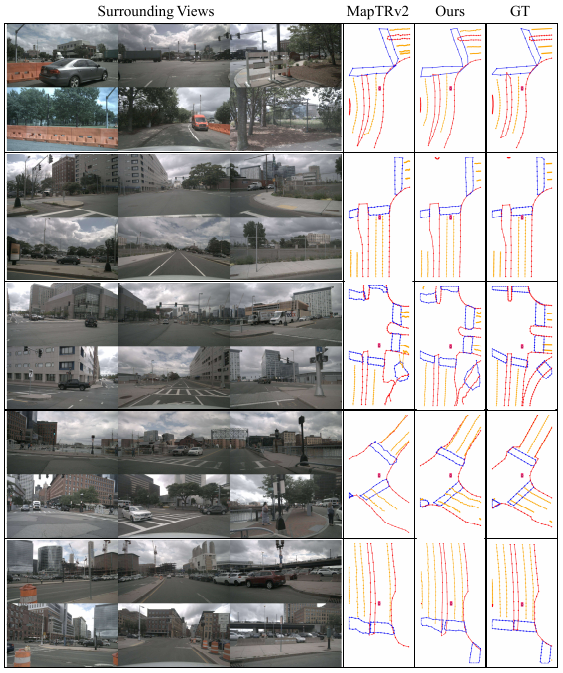}
\caption{Qualitative result comparison under the weather condition of cloudy on nuScenes.}
\label{fig:cloudy}
\end{figure*}

\begin{figure*}[h]
\centering
\includegraphics[width=0.8\textwidth]{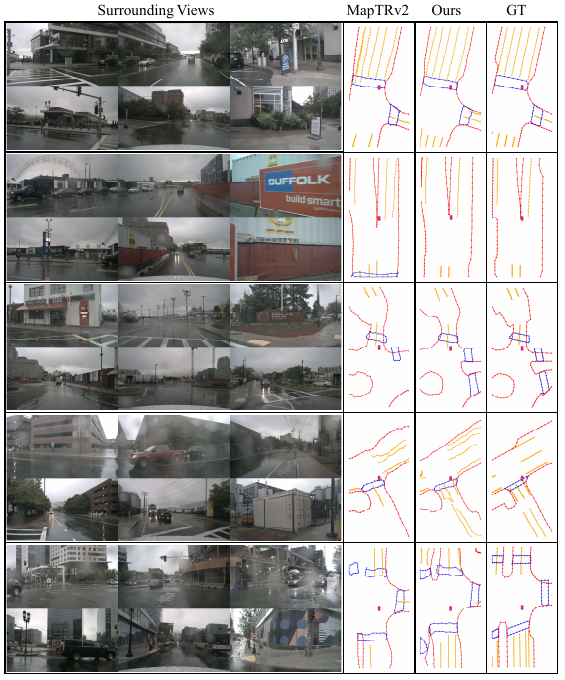}
\caption{Qualitative result comparison under the weather condition of rainy on nuScenes.}
\label{fig:rainy}
\end{figure*}

\begin{figure*}[h]
\centering
\includegraphics[width=0.8\textwidth]{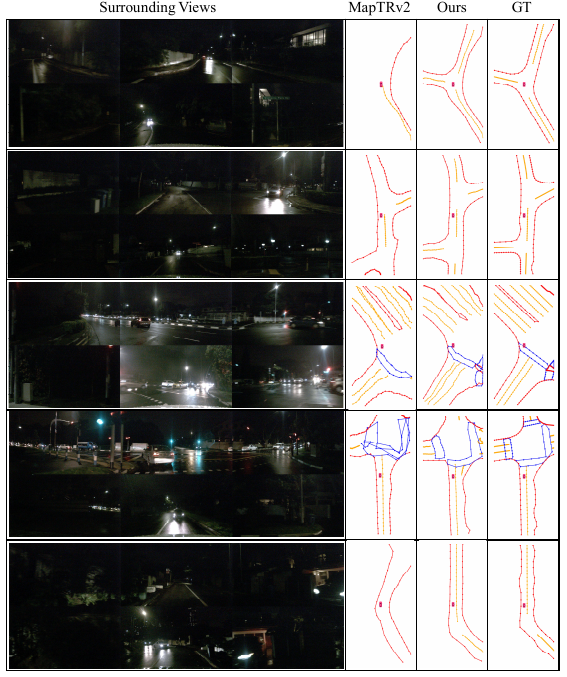}
\caption{Qualitative result comparison under the night condition on nuScenes.}
\label{fig:night}
\end{figure*}

\begin{figure*}[h]
\centering
\includegraphics[width=0.7\textwidth]{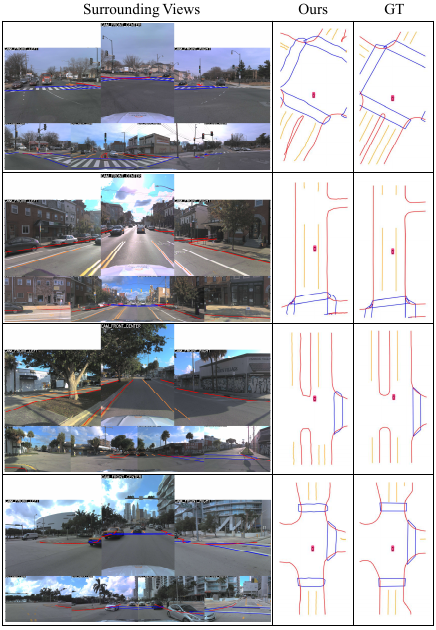}
\caption{Qualitative results on Argoverse2 val set. Since we generate a 3D vectorized HD map, the predictions can be rendered precisely on the surrounding view images.}
\label{fig:av2}
\end{figure*}

\begin{figure*}[h]
\centering
\includegraphics[width=0.7\textwidth]{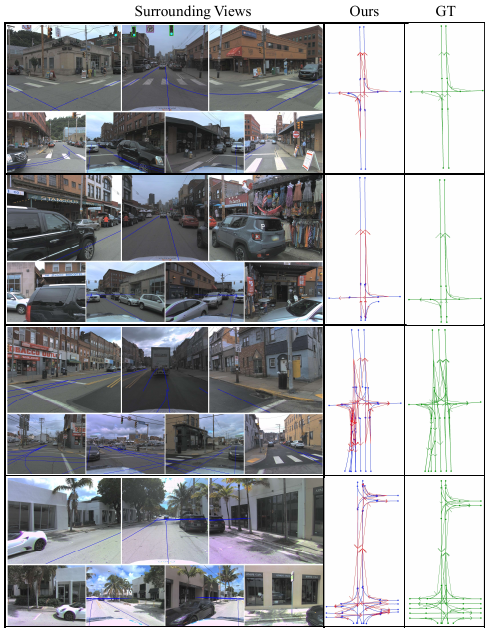}
\caption{Qualitative results on OpenLane-V2. The red arrows represent the predicted lane topology.}
\label{fig:openlanev2}
\end{figure*}

We visualize map construction results of PriorDrive under various weather conditions on nuScenes in \Cref{fig:scenarios}. PriorDrive maintains stable and impressive results.
Moreover, in \Cref{fig:cloudy}, \Cref{fig:rainy} and \Cref{fig:night}, we show the result comparison between MapTRv2 and the proposed PriorDrive on the nuScenes.  Finally, we also present the visualizations of Argoverse 2 in \Cref{fig:av2} and OpenLane-V2 in \Cref{fig:openlanev2}. Our results have more complete map representations, more accurate shape and point positions of map elements.

\end{document}